\documentclass[11pt]{article}

\usepackage[final]{acl}

\usepackage{times}
\usepackage{latexsym}
\usepackage[T1]{fontenc}
\usepackage[utf8]{inputenc}
\usepackage{microtype}
\usepackage{inconsolata}
\usepackage{graphicx}

\usepackage{hyperref}       
\usepackage{url}            
\usepackage{booktabs}       
\usepackage{amsfonts}       
\usepackage{nicefrac}       
\usepackage{microtype}      
\usepackage{xcolor}         
\usepackage[table]{xcolor}
\usepackage{wrapfig}
\usepackage{subcaption}
\usepackage{threeparttable}
\usepackage{amsmath}
\usepackage{amsthm}
\usepackage{paralist}
\usepackage{listings}
\usepackage[linesnumbered,ruled,vlined]{algorithm2e}
\usepackage[most]{tcolorbox}
\usepackage[framemethod=TikZ]{mdframed}
\usepackage{wrapfig}
\usepackage[linesnumbered,ruled,vlined]{algorithm2e}
\usepackage{multirow}

\usepackage{CJKutf8}

\newcommand{\floor}[1]{\lfloor #1 \rfloor}

\newcommand{\mycomment}[1]{}

\newcommand{\myparagraph}[1]{\vspace{1ex}\noindent\underline{\it #1.}\xspace}

\definecolor{lightred}{RGB}{255, 200, 200}
\definecolor{lightgreen}{RGB}{200, 255, 200}
\definecolor{deepviolet}{RGB}{96, 0, 96}

\lstdefinestyle{mypython}{
  language=Python,
  numberstyle=\tiny,
  frame=single,
  breaklines=true,
  basicstyle=\ttfamily\tiny,
  tabsize=4,
  showstringspaces=false,
}

\tcbset{
  promptbox/.style={
    top=10pt,
    colback=white,
    colframe=violet,
    colbacktitle=deepviolet,
    fontupper=\itshape,
    enhanced,
    center,
    attach boxed title to top left={yshift=-0.1in,xshift=0.15in},
    boxed title style={boxrule=0pt,colframe=white},
  }
}
\newtcolorbox{prompt}[2][]{promptbox,title=#2,#1}

\mdfsetup{%
  font=\small,
  innertopmargin=2pt,
  innerbottommargin=2pt,
  innerleftmargin=8pt,
  innerrightmargin=8pt,
  frametitleaboveskip=2pt,
  frametitlebelowskip=1pt,
  frametitlealignment=\center,
  middlelinecolor=blue
}
{
  \vspace{1ex}
  \begin{mdframed}
  \itshape
}
{
  \end{mdframed}
  \vspace{1ex}
}

\newcommand{\topk}{top-$k$\xspace}

\newcommand{\topp}{top-$p$\xspace}

\newcommand{\obs}{\text{obs}\xspace}

\newcommand{\aslgf}{ASL\_2pass\xspace}

\SetFuncSty{text}
\SetArgSty{text}
\SetKw{OR}{or}
\SetKw{AND}{and}
\SetKw{NOT}{not}
\SetKw{TRUE}{true}
\SetKw{FALSE}{false}
\SetKw{NULL}{nil}
\SetKw{Break}{break}
\SetKw{Continue}{continue}
\SetKwInOut{Input}{Input}
\SetKwInOut{Output}{Output}
\input{savespace}  

\title{Adaptive Layer Selection for Layer-Wise Token Pruning in LLM Inference}

\newcommand{\nec}{$^{1}$}
\newcommand{\inits}{$^{2}$}
\newcommand{\osaka}{$^{3}$}
\newcommand{\nagoya}{$^{4}$}
\newcommand{\osanag}{$^{3,4}$}

\author{\nec Rei Taniguchi\thanks{Work done at Osaka University.}, \inits Yuyang Dong, \osaka Makoto Onizuka, \osanag Chuan Xiao\\
\small{\nec NEC Corporation, \inits Initial S, \osaka Osaka University, \nagoya Nagoya University}\\
\small{rei-taniguchi@nec.com, \{onizuka, chuanx\}@ist.osaka-u.ac.jp, dongyuyang@initial-s.com}
}

\begin{document}

\pagestyle{plain}
\pagenumbering{arabic}

\maketitle

\def\thefootnote{}\footnotetext{Source code is available at \url{https://github.com/TANIGUCHIREI/ASL}.}
\def\thefootnote{\arabic{footnote}}

\begin{abstract}
  Due to the prevalence of large language models (LLMs), key-value (KV) cache reduction for LLM inference has received remarkable attention. 
  Among numerous works that have been proposed in recent years, layer-wise token pruning approaches, which select a subset of tokens at particular layers to retain in KV cache and prune others, are one of the most popular schemes. 
  They primarily adopt a set of pre-defined layers, at which tokens are selected. 
  Such design is inflexible in the sense that the accuracy significantly varies across tasks and deteriorates in harder tasks such as KV retrieval. 
  In this paper, we propose ASL, a training-free method that adaptively chooses the selection layer for KV cache reduction, exploiting the variance of token ranks ordered by attention score. 
  The proposed method balances the performance across different tasks while meeting the user-specified KV budget requirement. 
  ASL operates during the prefilling stage and can be jointly used with existing KV cache reduction methods such as SnapKV to optimize the decoding stage.
  By evaluations on the InfiniteBench, RULER, and NIAH benchmarks, we show that ASL, equipped with one-shot token selection,  adaptively trades inference speed for accuracy, outperforming state-of-the-art layer-wise token pruning methods in difficult tasks.
\end{abstract}

\section{Introduction}
\label{sec:intro}
Large language models (LLMs) have demonstrated remarkable capabilities in processing long contexts, enabling applications such as long document analysis, extended conversations, and software engineering. 
However, the memory footprint of LLM inference is a critical issue due to the key-value (KV) cache, which stores past tokens' key and value vectors for efficient generation. 

To address this challenge, numerous KV cache reduction techniques have emerged~\citep{zhang2023h2o,ge2023model,li2024snapkv,feng2024ada,fu2024not}. 
Among these, layer-wise token pruning methods~\citep{shi2024discovering,jo2025fastkv,cai2024pyramidkv,yang2024pyramidinfer,fu2024lazyllm} have recently gained considerable attention by exploiting attention patterns across Transformer layers~\citep{vaswani2017attention}. 
To achieve significant memory reduction while reducing accuracy loss, they select a subset of important tokens at particular layers, calculating attention only for these tokens and retaining their KV cache in subsequent layers. 

\begin{figure} [b]
    \centering
    \includegraphics[width=.9\linewidth]{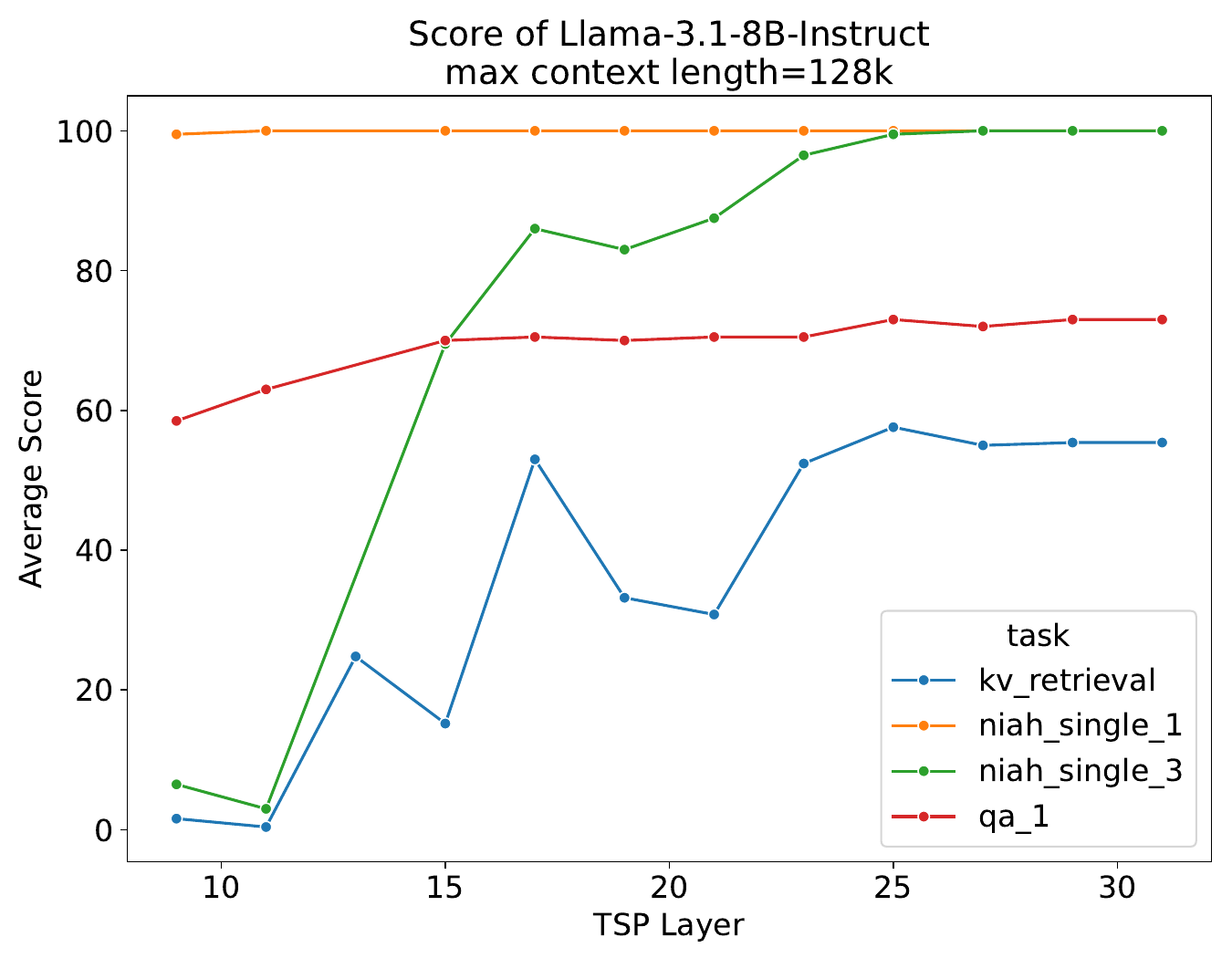}
    \caption{Performance of FastKV~\citep{jo2025fastkv} on four tasks under different selection layer (referred to as ``TSP layer'' in \citet{jo2025fastkv}) settings: 
    KV retrieval in InfiniteBench~\citep{zhang2024bench}, single-key NIAH (with varying difficulties) and QA in RULER~\citep{hsieh2024ruler}. 
    KV compression before selection layer is disabled to highlight the impact.}
    \label{fig:fastkv-performance-varying-tsp}
\end{figure}

\begin{figure*}[t]
    \centering
    \includegraphics[width=\linewidth]{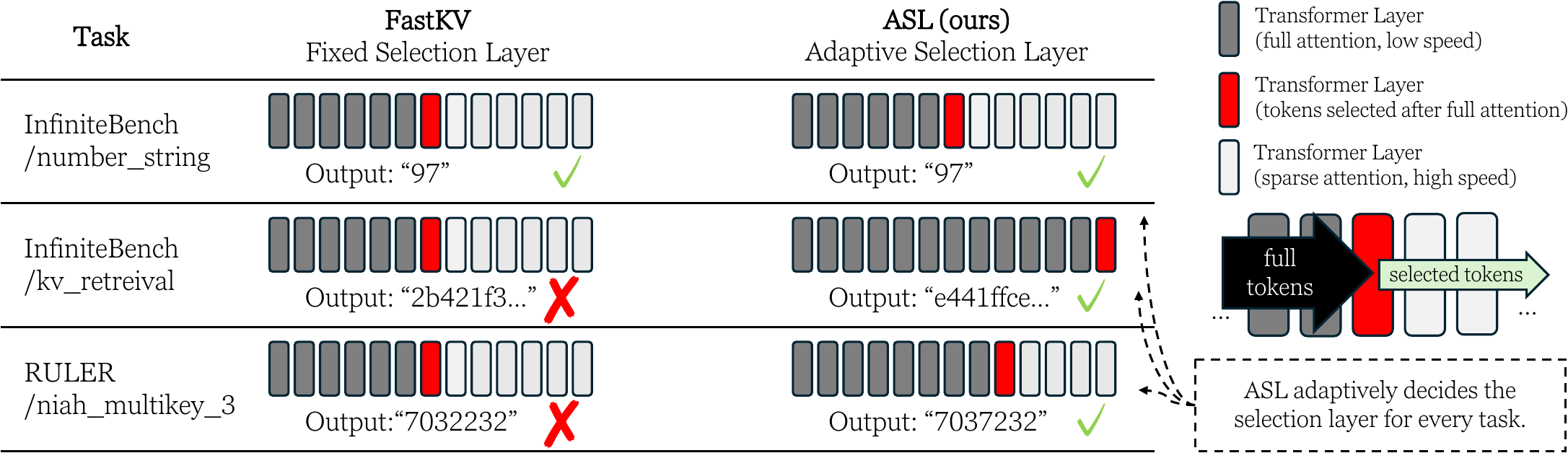}
    \caption{Comparison of FastKV and ASL.}
    \label{fig:fastkv-vs-asl}
\end{figure*}

Despite their effectiveness, existing layer-wise token pruning methods suffer from a critical limitation: tokens are selected on pre-defined, fixed layers (referred to as ``selection layers'') that are determined independently of the task. 
As illustrated in Figure~\ref{fig:fastkv-performance-varying-tsp}, this inflexible design leads to substantial performance variation across tasks. 
For simpler tasks like question answering (QA) where relevant information can be identified early, token selection at middle layers (e.g., layer 15 for Llama 3.1 8B, as suggested in \citet{jo2025fastkv}) achieves strong performance. 
However, for harder tasks like KV retrieval, where high semantic similarity between context and query makes early identification difficult, the same selection layers result in severe accuracy degradation. 
To maintain acceptable performance on challenging tasks, these methods must either postpone token selection to deeper layers or increase the KV budget, thereby compromising their memory reduction benefits.


In this paper, we propose Adaptive Selection Layer (ASL), a task-aware method that adaptively determines the selection layer based on the attention patterns observed during inference. 
Our idea is to monitor the variance of token ranks ordered by attention score across consecutive layers. 
When this variance decreases below a threshold, it signals that attention scores have consistently focused on a stable subset of tokens, indicating the moment for token selection. 
By computing variance and comparing it against a user-specified threshold, ASL adapts to tasks of varying difficulty without requiring manual tuning for each scenario.
Figure~\ref{fig:fastkv-vs-asl} depicts the comparison of FastKV and ASL.

ASL operates during the prefilling stage with minimal overhead, storing only pooled attention scores from recent layers. 
Once the selection layer is determined, tokens are selected and propagated to all subsequent layers in a one-shot manner.
To meet specific KV budget requirements, ASL can be seamlessly integrated with existing methods. 
For example, it can be combined with SnapKV~\citep{li2024snapkv} to optimize the decoding stage, and can be integrated into GemFilter~\citep{shi2024discovering} to enable a two-pass strategy for enhanced accuracy.

We evaluate ASL on three long-context benchmarks, InfiniteBench, RULER, and Needle in a Haystack (NIAH)~\citep{niah}, with up to 256k context lengths. 
Experimental results demonstrate that ASL can outperform state-of-the-art layer-wise token selection methods including FastKV, GemFilter, and PyramidInfer in accuracy, while maintaining comparable decoding speed and KV cache reduction. 

Our contributions are summarized as follows:
\begin{inparaenum} [(1)]
  \item We identify the inflexibility of fixed selection layers in existing layer-wise pruning methods and demonstrate how task difficulty affects optimal layer selection.
  \item We propose ASL, a novel adaptive method that determines selection layers by monitoring the variance of token ranks ordered by attention score across layers, enabling task-aware KV cache reduction.
  \item We demonstrate that ASL can be integrated with existing KV cache reduction methods to optimize both prefilling and decoding stages while meeting user-specified KV budgets.
  \item Through comprehensive experiments on multiple benchmarks, we show that ASL achieves superior accuracy-efficiency trade-offs compared to state-of-the-art methods, particularly excelling on difficult tasks where existing methods fail.
\end{inparaenum}
\section{Preliminaries}
\label{sec:prelim}
LLM inference typically involves two stages:
\begin{itemize}
  \item \textbf{Prefilling:} This stage occurs when the LLM processes the input prompt at once. 
  For each Transformer layer, the LLM computes query, 
  key, 
  and value 
  vectors for all tokens. 
  Attention is applied across all pairs of input tokens, leading to full self-attention.  
  KV cache is initialized and stores keys and values for all past tokens. 
  \item \textbf{Decoding:} This stage outputs tokens one at a time, using the previously generated tokens and their cached keys and values. 
  Only the latest token is passed through the Transformer. 
  The LLM computes the query for this new token, and calculates attention between the new query and all cached keys and values from prior tokens, which are retrieved from the KV cache. 
\end{itemize}

\begin{table*} [t]
    \small
    \caption{List of notable layer-wise token pruning methods. Taxonomy is explained in Section~\ref{sec:layer-wise-token-pruning}.}
    \centering
    \begin{tabular}{c | c}
        \toprule
        \rowcolor{gray!20}
        \textbf{Type} & \textbf{Methods} \\
        \midrule
        One-shot      & GemFilter~\citep{shi2024discovering}, FastKV~\citep{jo2025fastkv} \\ \hline
        \multirow{2}{*}{Progressive}   & PyramidKV~\citep{cai2024pyramidkv}, PyramidInfer~\citep{yang2024pyramidinfer}, LazyLLM~\citep{fu2024lazyllm}, \\ 
                & PromptDistill~\citep{jin2025promptdistill}, SlimInfer~\citep{long2025sliminfer} \\ \hline
        Sandwiched    & OmniKV~\citep{hao2025omnikv} \\ \hline 
        Grouped       & SqueezeAttention~\citep{wang2024squeezeattention}, EvolKV~\citep{yu2025evolkv} \\ \hline
        Adaptive      & DynamicKV~\citep{zhou2024dynamickv}, CAKE~\citep{qin2025cake} \\
        \bottomrule
    \end{tabular}
    \label{tab:layer-wise-methods}
\end{table*}

\subsection{KV Cache Reduction}
To reduce KV cache size, which is essential in long-context scenarios, token eviction techniques have been extensively explored. 
StreamingLLM~\citep{xiao2023efficient} keeps only global (first few tokens, a.k.a. attention sinks) and local (most recent) tokens in the sequence. 
H2O~\citep{zhang2023h2o}, FastGen~\citep{ge2023model}, and SnapKV~\citep{li2024snapkv} exploit attention sparsity and retain the most influential tokens' KV cache for the decoding stage, which are decided using heuristics during the prefilling stage. AdaKV~\citep{feng2024ada} and HeadKV~\citep{fu2024not} extend SnapKV with a head-wise budget allocation strategy. 

Another line of works does not evict tokens but loads a subset of tokens' KV cache by leveraging attention sparsity~\citep{tang2024quest,singhania2024loki} or offloading to CPU memory~\citep{lee2024infinigen,sun2024shadowkv}. Besides these methods based on token selection, quantization~\citep{liu2024kivi,hooper2024kvquant} is also an approach to reducing KV cache size. Moreover, LSH~\citep{charikar2002similarity} has been utilized to approximate the attention score distribution and estimate attention output~\citep{chen2024magicpig}. For other works on KV cache reduction, we refer readers to a GitHub repository~\citep{awesome} for an up-to-date list of papers. 

\subsection{Layer-Wise Token Pruning}
\label{sec:layer-wise-token-pruning}
Among the methods for KV cache reduction, many recent ones adopt a layer-wise token pruning paradigm. Observing the attention patterns across Transformer layers, they exploit either the similarity between adjacent layers or dissimilarity in earlier and later layers, and select a subset of tokens---typically by \topk or \topp (cumulative attention score is no less than $p$ percentile of full attention)---to calculate attention and retain in the KV cache. Other tokens are pruned or offloaded to CPU memory~\citep{hao2025omnikv,long2025sliminfer}. 

Table~\ref{tab:layer-wise-methods} summarizes a list of notable layer-wise token pruning methods, categorized into five types. 
\begin{itemize}
    \item \textbf{One-shot} methods select tokens only once at a specific layer, and all deeper layers process only the selected tokens. FastKV~\citep{jo2025fastkv} and GemFilter~\citep{shi2024discovering} are two representative methods in this category. FastKV adopts a one-pass strategy: (1) from layer 0 to the selection layer, full attention is calculated; (2) at the selection layer, \topk tokens are selected; (3) for subsequent layers, attention is calculated only for the selected tokens. To meet the KV budget requirement, SnapKV~\citep{li2024snapkv} is jointly used in FastKV to compress the KV cache from layer 0 to the selection layer. GemFilter adopts a two-pass strategy: (1) the first pass calculates full attention and selects \topk tokens at the selection layer; (2) the second pass starts from layer 0 and processes all layers, with attention calculated and KV cache retained for the selected tokens only. 
    \item \textbf{Progressive} methods employ a multi-shot token selection scheme, progressively reducing the tokens processed for attention and retained in the KV cache. Most methods in this category, except PyramidKV~\citep{cai2024pyramidkv}, are monotonic methods---a token pruned at a layer is also pruned at all later layers. 
    \item \textbf{Sandwiched} methods (in particular, OmniKV~\citep{hao2025omnikv}) calculate attention for all tokens at some layers and selected tokens for others, thereby exhibiting a sandwich shape across layers. 
    \item \textbf{Grouped} methods divide layers into several groups and allocate a KV cache budget to each group, such that the allocated values sum up to a user-specified total budget. The allocation is processed online (i.e., during inference) in SqueezeAttention~\citep{wang2024squeezeattention} and offline in EvolKV~\citep{yu2025evolkv}. 
    \item \textbf{Adaptive} methods directly allocate the KV cache budget to each layer. Online allocation is used in DynamicKV~\citep{zhou2024dynamickv} and CAKE~\citep{qin2025cake}. 
\end{itemize}


In addition to token pruning, layer-wise observations have also been used to build cross-layer merging or sharing methods, such as block-based layer pruning (BBLP)~\citep{gromov2024unreasonable}, MiniCache~\citep{liu2024minicache}, FoldGPT~\citep{liu2024foldgpt}, SwiftKV~\citep{qiao2024swiftkv}, where a small amount of fine-tuning or distillation is needed in BBLP, FoldGPT, and SwiftKV. It is also noteworthy to mention that layer-wise behavior is evaluated in previous studies such as \citet{xiao2023efficient} and \citet{li2024snapkv}, yet they do not belong to the category of layer-wise token pruning because their token selection in each layer is performed independently. 






\section{Observations}
\label{sec:obs}
The layer-wise token pruning methods summarized in Table~\ref{tab:layer-wise-methods}, despite achieving significant peak memory reduction and fast decoding speed, only a minority of them, including GemFilter, FastKV, PyramidInfer, LazyLLM, and SlimInfer, optimize time to first token (TTFT), a key metric that evaluates the efficiency of the prefilling stage. 
They select tokens either at a pre-defined set of layers (GemFilter, FastKV, LazyLLM, and SlimInfer) or with a decay ratio across layers (PyramidInfer). 

Such design is inflexible in the sense that the difficulty of tasks is not considered, rendering these methods either incapable of delivering competitive performance for harder tasks such as KV retrieval and multi-key NIAH, or have to compromise its KV cache reduction (by postponing token selection to later layers or increasing the KV budget) to cope with these harder tasks, as we have seen in Figure~\ref{fig:fastkv-performance-varying-tsp}.

As discussed in \citet{minference}, the difficulty of the task originates from the semantic similarity between the question and the context. A task tends to be easier if the similarity is low, as the LLM can easily identify the context related to the question. In such tasks, it is possible to locate the tokens necessary for the answer in early layers. In contrast, in the harder KV retrieval task, the context consists of key-value pairs, rendering high similarity between the question and the context. As a result, token selection at early layers cannot successfully locate the tokens required for the answer. 

\begin{figure} [t]
    \centering
    \includegraphics[width=\linewidth]{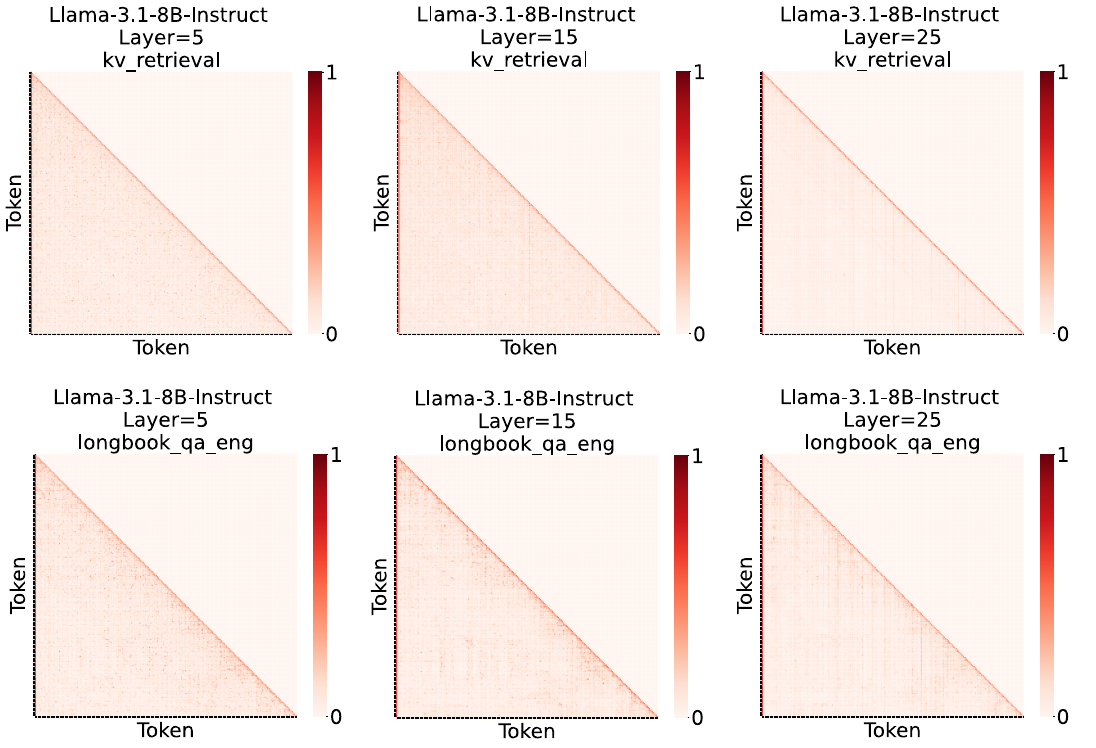}
    \caption{Attention patterns of KV retrieval (upper) and QA (lower). At early layers, the attention scores are roughly uniformly distributed across the context. At middle layers, a subset of tokens exhibit high scores (as shown in stripe-like regions). The scores are more localized at deep layers, focusing to a smaller subset of tokens (as shown in thin vertical lines).}
    \label{fig:attn-pttns}
\end{figure}

\citet{cai2024pyramidkv} found that in early layers, attention scores are generally distributed in a uniform manner across the tokens in the context, and in later layers, the scores tend to localize to a fixed subset of tokens. While such patterns were observed for an RAG task in \citet{cai2024pyramidkv}, we find that they apply to various tasks. Figure~\ref{fig:attn-pttns} shows that both KV retrieval and QA tasks, whose difficulties significantly differ, exhibit similar attention patterns. Based on this observation, we can design an adaptive method that decides the selection layer for layer-wise token pruning: token pruning is performed when attention scores start to focus consistently on a small subset of tokens. 

\begin{figure}[t]
    \small
    \centering
    \begin{subfigure}{\linewidth}
        \centering
        \includegraphics[width=\linewidth]{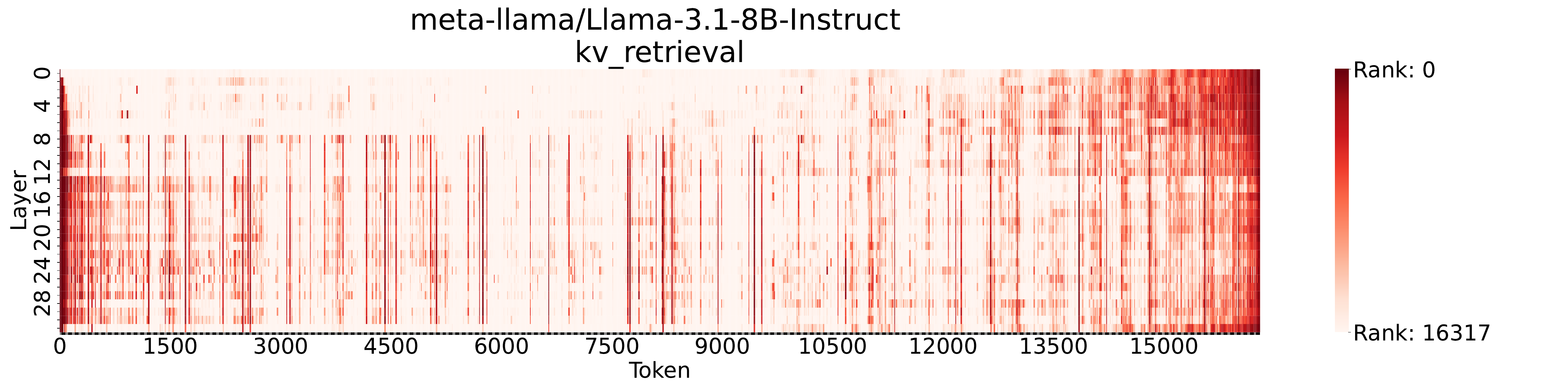}
        \label{fig:rank-pttn-kv-retrieval}
    \end{subfigure}
    \begin{subfigure}{\linewidth}
        \centering
        \includegraphics[width=\linewidth]{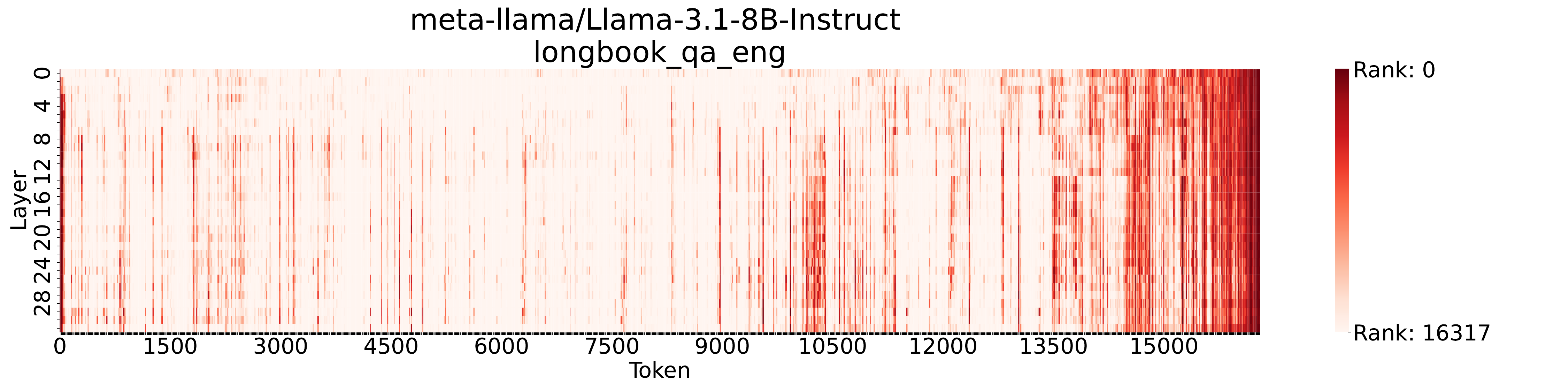}
        \label{fig:rank-pttn-longbook-qa-eng}
    \end{subfigure}
    \caption{Ranks patterns across layers for KV retrieval (upper) and QA (lower), context length = 16k.}
    \label{fig:rank-pttn}
\end{figure}

\section{Adaptive Selection Layer}
\label{sec:asl}

\begin{figure*}[t]
    \centering
    \includegraphics[width=.9\linewidth]{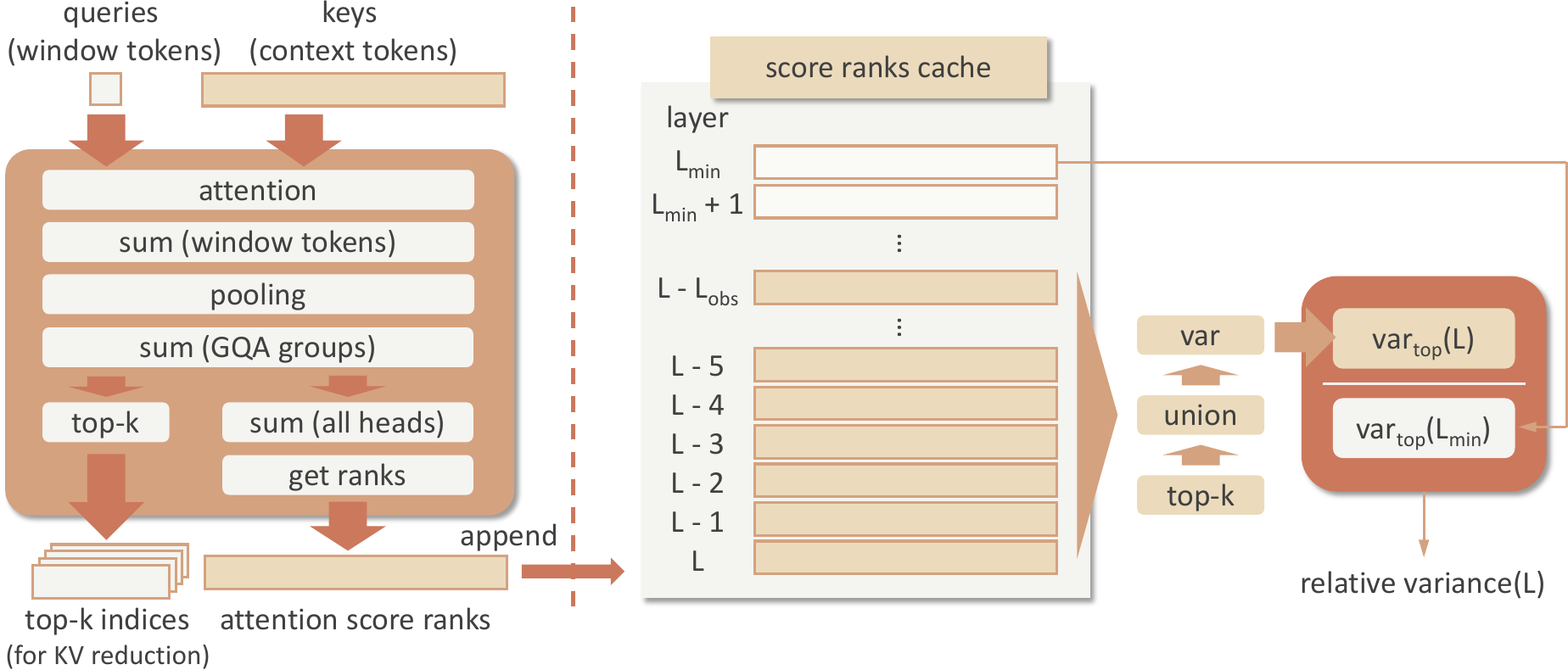}
    \caption{Relative variance calculation.}
    \label{fig:relative-variance-calculation}
\end{figure*}

\begin{figure}[t]
    \includegraphics[width=\linewidth]{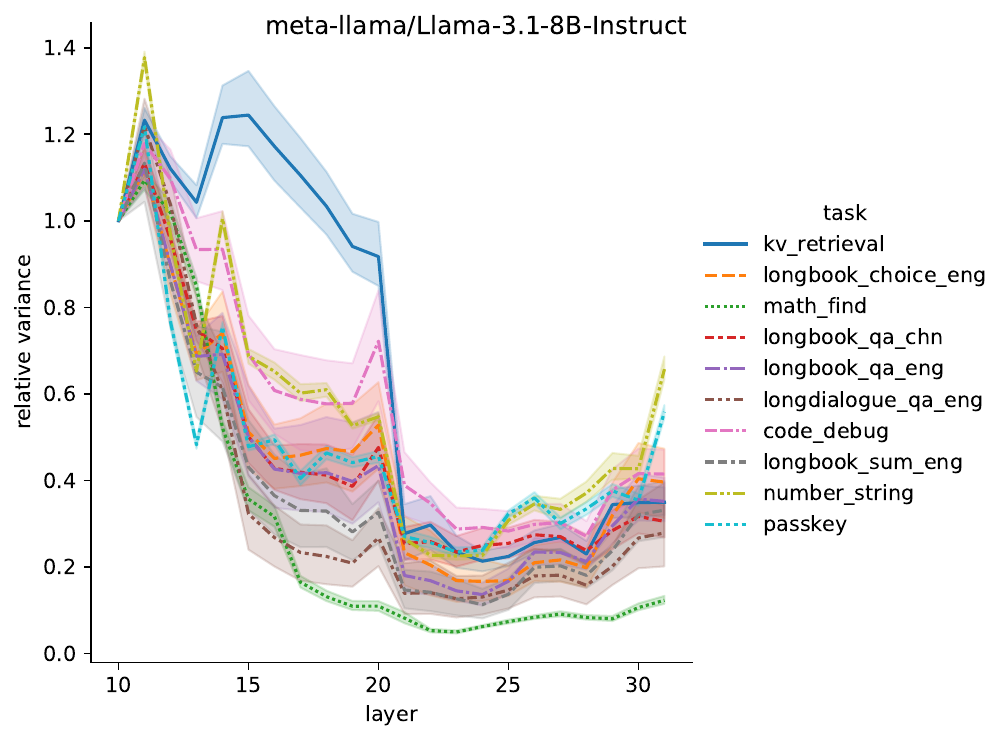}
    \caption{Relative variances across layers for 10 tasks in the InfiniteBench benchmark.}
    \label{fig:task-relative-variance}
\end{figure}

\subsection{Variance of Token Ranks}
\label{sec:variance}
To find when attention scores start to focus consistently to a subset of tokens, our idea is to monitor how the tokens in the context, ranked by the attention score, vary across layers. 
As shown in Figure~\ref{fig:rank-pttn}, the top ranks tend to fix at deeper layers for both KV retrieval and QA tasks. 
Seeing this, we calculate the variance of ranks as an indicator: 
a small variance of ranks indicates not only the scores are more focused to a fixed subset of tokens, but also the order of these tokens, ranked by the score, is relatively fixed. 

Figure~\ref{fig:relative-variance-calculation} depicts the method that calculates the variance. We start with layer $L_{\min}$ of the LLM and consider the ranks in every $L_{\obs}$ consecutive layers, thereby eliminating the effect of the layers too early or too distant from the current layer. 
$L_{\min}$ and $L_{\obs}$ and two hyperparameters. 
Attention scores are aggregated over attention heads. 
Instead of using raw attention scores, we perform 1D average pooling to smooth noise and capture contiguous regions in the context, in line with SnapKV~\citep{li2024snapkv}. 
The pooled attention scores, denoted $PA(L)$, are calculated as follows.
\begin{align*}
  \text{PA} = \text{pool}\left( 
  \text{softmax}\left(
  \frac{\mathbf{q}_{w} \mathbf{k}_{c} + \mathbf{m}_{w}}{\sqrt{d}}
  \right)
  \right)
\end{align*}
where $\mathbf{q}_{w}$ denotes the query vector of current window, $\mathbf{k}_{c}$ denotes the key vector of the context, and $\mathbf{m}_{w}$ denotes causal masking. 

We sum up the average pooled attention scores over GQA head groups, which are used to order the tokens. 
\begin{align*}
  \text{scores} = \sum_{i \in \text{groups}} \sum_{j \in \text{heads}} \text{PA}[i,j,:].
\end{align*}

Due to the attention sparsity, not all tokens in the context need to be considered for variance calculation. 
From layers $L - L_{\obs}$ to $L$, we identify the \topk tokens and get a union of the \topk's: 
\begin{align*}
  \text{top}(L) = \text{sort} \left( \bigcup_{l = L - L_{\obs}}^{L} \text{top-$k$}(\text{scores}(l)) \right).
\end{align*}

The union serves as a subset of tokens with high attention scores. The variance of the ranks is then calculated for the tokens in the union: 
\begin{align*}
  \text{var}_{\text{top}}(L) = 
  \frac{1}{|\text{top}(L)|} \sum_{t \in \text{top}(L)} \text{var} \left( R_t[L - L_{\obs}, L] \right)
\end{align*}
where $R_t[L - L_{\obs}, L]$ denotes the set of token $t$'s ranks from layers $L - L_{\obs}$ to $L$. 

As a task-aware design, we divide each variance by the initial variance at layer $L_{\min}$ to obtain a relative variance 
\begin{align*}
  \text{relative\_variance}(L) = \frac{\text{var}_{\text{top}}(L)}{\text{var}_{\text{top}}(L_{\text{min}})}. 
\end{align*}
The relative variance is then compared with a user-specified threshold $\tau$. 
If the relative variance is less than $\tau$, the attention scores are regarded as consistently focused, and the current layer becomes the selection layer.

Figure~\ref{fig:task-relative-variance} shows how the relative variance changes with layers for 10 tasks in the InfiniteBench benchmark. 
If we set $\tau = 0.4$, the math find and dialog QA tasks will have the earliest selection layer among the 10 tasks, while the KV retrieval and code debug tasks have the deepest selection layer. 

We name this method adaptive selection layer (ASL) and provide pseudo-codes in Algorithms~\ref{alg:update-kv-rank} and~\ref{alg:select-selection-layer}, Appendix~\ref{sec:exp-setup-detail}, both acting in the prefilling stage. 







\subsection{Using ASL for LLM Inference}
\label{sec:inference}

\myparagraph{Prefilling}
During the prefilling stage, from layer $L_{\min}$, ASL computes and stores the pooled mean attention scores, which are used in the next $L_{\obs}$ layers. 
Since the pooled scores are aggregated over the attention heads and only the most recent $L_{\obs}$ layers need to be kept, they incur only a small amount of peak memory usage. 
For example, when $L_{\obs} = 8$, for Llama 3.1 8B with 32 layers and 8 KV heads, the pooled scores incurs an additional $\frac{L_{\obs}}{32 \times 8} = \frac{1}{32}$ memory usage compared to attention calculation.
When the selection layer is determined, token selection is performed and the prefilling proceeds with only the selected tokens. 

\myparagraph{Decoding}
When new tokens are decoded, only the KV entries for these new tokens are added to the cache and used in subsequent attention calculation. 

To meet the KV budget requirement, ASL can work jointly with existing KV cache reduction methods. For example, SnapKV can be used to select KV entries for all the layers prior to the selection layer. Supposing the same $k$ is used for the \topk selection by ASL and SnapKV, this integration guarantees that the KV cache size of each layer is exactly $k$ during decoding. 
ASL can be also used with the two-pass method GemFilter. 
In the first pass, tokens are selected once the selection layer is determined. 
Then, we start the second pass from layer 0 by processing only the selected tokens.





We provide a theoretical analysis in Appendix~\ref{sec:theoretical-analysis} for the costs of ASL's prefilling and decoding, comparing with FastKV and full attention.
\begin{table*}[t]
    \small
    \centering
    \setlength{\tabcolsep}{1ex}
    \caption{Accuracy ($\uparrow$) comparison on InfiniteBench. PyramidInfer encounters OOM for all tasks and is not reported.}
    \resizebox{\textwidth}{!}{
    \begin{tabular}{l|cccccccccc|c}
    \toprule
    \textbf{Methods}       & \textbf{En.Sum} & \textbf{En.QA} & \textbf{En.MC} & \textbf{En.Dia} & \textbf{Zh.QA} & \textbf{Code.Debug} & \textbf{Math.Find} & \textbf{Retr.PassKey} & \textbf{Retr.Num} & \textbf{Retr.KV} & \textbf{Avg.} \\
    \midrule
    \midrule
    \multicolumn{12}{c}{\textbf{Llama-3.1-8B-UL, KV Budget = Full.}} \\
    \midrule
    Full KV       & 27.2 & 17.73 & 65.5 & 11.5 & 21.26 & 0 & 38 & 100 & 99.32 & 16.2 & 39.7 \\
    \midrule
    \multicolumn{12}{c}{\textbf{Llama-3.1-8B-UL, KV Budget = 2048.}} \\
    \midrule
    SnapKV        & 21.86 & 17.8 & 65.5 & 8.5 & 20.71 & 0 & 36.29 & 100 & 98.47 & 2 & 37.1 \\
    FastKV        & 21.63 & 17.33 & 67.69 & 6 & 20.14 & 0 & 32.29 & 100 & 98.47 & 0.6 & 36.4 \\
    \rowcolor[HTML]{DAE8FC}
    ASL           & 21.26 & 18.42 & 65.5 & 5.5 & 20.84 & 0 & 35.43 & 100 & 98.47 & 1.8 & 36.7 \\
    GemFilter     & 5.72 & 15.78 & 53.71 & 13.5 & 18.88 & 25.89 & 36.29 & 100 & 100 & 0 & 37 \\
    \rowcolor[HTML]{DAE8FC}
    \aslgf        & 5.81 & 18.37 & 64.63 & 12 & 22.3 & 25.38 & 27.71 & 100 & 100 & 2.2 & \textbf{37.8} \\
    \midrule
    \multicolumn{12}{c}{\textbf{Llama-3.1-8B-UL, KV Budget = Full (before selection) \& 2048 (after selection).}} \\
    \midrule
    FastKV        & 22.86 & 16.33 & 67.69 & 6.5 & 20.23 & 0 & 32 & 100 & 99.49 & 3.2 & 36.8 \\
    \rowcolor[HTML]{DAE8FC}
    ASL           & 24.54 & 17.82 & 65.5 & 7 & 21.66 & 0 & 35.71 & 100 & 99.49 & 15.4 & \textbf{38.7} \\
    \midrule
    \midrule
    \multicolumn{12}{c}{\textbf{Qwen2.5-7B, KV Budget = Full.}} \\
    \midrule
    Full KV       & 33.42 & 12.63 & 69.43 & 9 & 12.76 & 0.76 & 38.57 & 99.83 & 100 & 66.4 & 44.3 \\
    \midrule
    \multicolumn{12}{c}{\textbf{Qwen2.5-7B, KV Budget = 2048.}} \\
    \midrule
    SnapKV        & 28.22 & 12.02 & 69.43 & 3.5 & 12.02 & 0.76 & 35.43 & 95.93 & 100 & 0.4 & 35.8 \\
    FastKV        & 27.95 & 11.05 & 68.56 & 6.5 & 11.01 & 0.51 & 34.86 & 99.49 & 100 & 1 & \textbf{36.1} \\
    \rowcolor[HTML]{DAE8FC}
    ASL           & 28.3 & 11.89 & 70.31 & 3 & 12 & 0.25 & 34.86 & 99.66 & 100 & 0.6 & \textbf{36.1} \\
    GemFilter     & 4.69 & 5.66 & 27.51 & 21.5 & 6.57 & 3.3 & 7.71 & 99.66 & 100 & 0 & 27.7 \\
    \rowcolor[HTML]{DAE8FC}
    \aslgf        & 4.81 & 12.98 & 63.32 & 13 & 11.38 & 3.3 & 13.71 & 98.47 & 100 & 5.2 & 32.6 \\
    \midrule
    \multicolumn{12}{c}{\textbf{Qwen2.5-7B, KV Budget = Full (before selection) \& 2048 (after selection).}} \\
    \midrule
    FastKV        & 27.99 & 12.15 & 68.56 & 5 & 11.72 & 0.51 & 35.14 & 99.83 & 100 & 1 & 36.2 \\
    \rowcolor[HTML]{DAE8FC}
    ASL           & 29.52 & 12.28 & 70.31 & 6 & 11.88 & 0.25 & 35.14 & 99.83 & 100 & 52.6 & \textbf{41.8} \\
    \bottomrule
    \end{tabular}
    }
    \label{tab:score-infinitebench}
\end{table*}

\section{Experiments}
\label{sec:exp}

\subsection{Experimental Setup}

\myparagraph{Models}
We evaluate two long-context LLMs: 
\begin{inparaenum} [(1)]
    \item Llama-3.1-Nemotron-8B-UltraLong-1M-Instruct (Llama-3.1-8B-UL, for short), with 32 layers, and 
    \item Qwen2.5-7B-Instruct-1M (Qwen2.5-7B, for short), with 28 layers. 
\end{inparaenum}

\myparagraph{Benchmarks}
We use three benchmarks for evaluation:
\begin{inparaenum} [(1)]
    \item InfiniBench~\citep{zhang2024bench}, with an average context length of 214k. 
    \item RULER~\citep{hsieh2024ruler}, with context length ranging from 4k to 128k, and 
    \item Needle in a Haystack (NIAH)~\citep{niah}, with context length from 1k to 256k. 
\end{inparaenum}

\myparagraph{Methods}
We compare ASL with three layer-wise token pruning methods: 
\begin{inparaenum} [(1)]
  \item FastKV~\citep{jo2025fastkv}, 
  \item GemFilter~\citep{shi2024discovering}, and 
  \item PyramidInfer~\citep{yang2024pyramidinfer}. 
\end{inparaenum}
As introduced in Section~\ref{sec:variance}, we select \topk tokens in ASL ranked by pooled mean scores and equip ASL with SnapKV~\citep{li2024snapkv} to optimize decoding. This method, referred to as ASL, serves as a one-pass solution. 
Moreover, we integrate ASL with GemFilter as a two-pass solution, referred to as \aslgf. 
We set the default KV budget size to 2048. 
Moreover, to show how ASL compares to FastKV on prefilling, we consider a setting with full KV cache before token selection and a KV budget of 2048 after token selection; i.e., SnapKV is not applied for ASL and KV compression is disabled before token selection in FastKV. 
$L_{\min}$ is 10 for Llama-3.1-8B-UL and 9 for Qwen2.5-7B. 
$L_{\obs}$ = 8. The default value of $\tau = 0.3$. 
Other details can be found in Appendix~\ref{sec:exp-setup-detail}.

\begin{table*}[t]
    \small
    \centering
    \setlength{\tabcolsep}{3ex}
    \caption{Accuracy ($\uparrow$) comparison on RULER, averaged over 13 tasks.}
    \begin{tabular}{l|cccccc}
    \toprule
    \textbf{Methods}       & \textbf{4k} & \textbf{8k} & \textbf{16k} & \textbf{32k} & \textbf{64k} & \textbf{128k} \\
    \midrule
    \midrule
    \multicolumn{7}{c}{\textbf{Llama-3.1-8B-UL, KV Budget = Full.}} \\
    \midrule
    Full KV       & 94.2 & 92.0 & 89.5 & 81.1 & 73.1 & 68.6 \\
    \midrule
    \multicolumn{7}{c}{\textbf{Llama-3.1-8B-UL, KV Budget = 2048.}} \\
    \midrule
    SnapKV        & \textbf{93.8} & \textbf{88.3} & \textbf{81.0} & 74.8 & 63.6 & 55.6 \\
    FastKV        & 93.4 & 85.8 & 79.3 & 69.5 & 60.1 & 55.1 \\
    PyramidInfer  & 76.4 & 74.7 & OOM & OOM & OOM & OOM \\
    \rowcolor[HTML]{DAE8FC}
    ASL           & 93.7 & 87.0 & 79.7 & 73.0 & 63.2 & 56.1 \\
    GemFilter     & 92.5 & 81.1 & 79.0 & 76.0 & \textbf{71.0} & \textbf{66.7} \\
    \rowcolor[HTML]{DAE8FC}
    \aslgf        & 82.2  & 74.6  & 74.8  & \textbf{78.5}  & 69.7  & 54.5 \\
    \midrule
    \multicolumn{7}{c}{\textbf{Llama-3.1-8B-UL, KV Budget = Full (before selection) \& 2048 (after selection).}} \\
    \midrule
    FastKV        & 93.7 & 87.3 & 82.9 & 73.0 & 63.4 & 60.6 \\
    \rowcolor[HTML]{DAE8FC}
    ASL           & \textbf{94.2} & \textbf{90.5} & \textbf{87.2} & \textbf{80.0} & \textbf{70.5} & \textbf{69.2} \\
    \midrule
    \midrule
    \multicolumn{7}{c}{\textbf{Qwen2.5-7B, KV Budget = Full.}} \\
    \midrule
    Full KV       & 94.0 & 93.0 & 92.7 & 89.6 & 86.5 & 82.3 \\
    \midrule
    \multicolumn{7}{c}{\textbf{Qwen2.5-7B, KV Budget = 2048.}} \\
    \midrule
    SnapKV        & 89.9 & 77.9 & 75.3 & 71.8 & 66.8 & 65.0 \\
    FastKV        & \textbf{91.0} & \textbf{79.7} & 75.3 & 69.9 & 63.2 & 59.1 \\
    PyramidInfer  & 37.9 & 30.2 & OOM  & OOM  & OOM  & OOM \\
    \rowcolor[HTML]{DAE8FC}
    ASL           & 89.9 & 77.5 & \textbf{75.5} & \textbf{75.1} & \textbf{71.7} & \textbf{66.4} \\
    GemFilter     & 79.7 & 67.7 & 66.9 & 63.8 & 60.4 & 56.7 \\
    \rowcolor[HTML]{DAE8FC}
    \aslgf        & 78.3 & 70.7 & 67.9 & 71.5 & 69.1 & 62.5 \\
    \midrule
    \multicolumn{7}{c}{\textbf{Qwen2.5-7B, KV Budget = Full (before selection) \& 2048 (after selection).}} \\
    \midrule
    FastKV        & 91.3 & 80.6 & 75.8 & 70.6 & 64.2 & 59.1 \\
    \rowcolor[HTML]{DAE8FC}
    ASL           & \textbf{94.0} & \textbf{91.2} & \textbf{88.0} & \textbf{83.6} & \textbf{80.9} & \textbf{74.2} \\
    \bottomrule
    \end{tabular}
    \label{tab:score-ruler}
\end{table*}
\begin{figure*}[t]
    \small
    \centering
    \begin{subfigure}{0.33\textwidth}
        \centering
        \includegraphics[width=\linewidth]{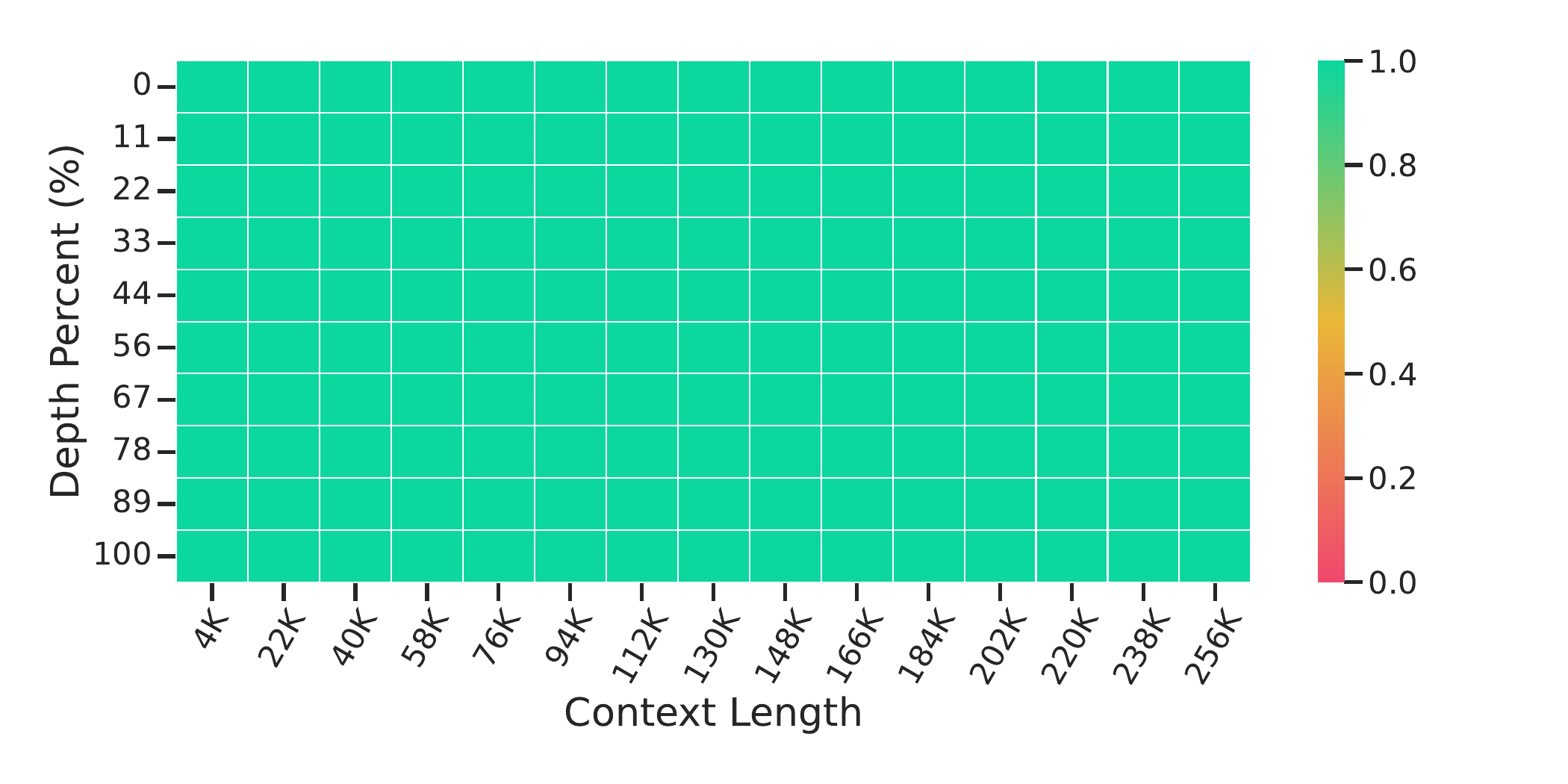}
        \caption{Full KV.}
        \label{fig:niah-results-qwen257b-fullkv}
    \end{subfigure}
    \begin{subfigure}{0.33\textwidth}
        \centering
        \includegraphics[width=\linewidth]{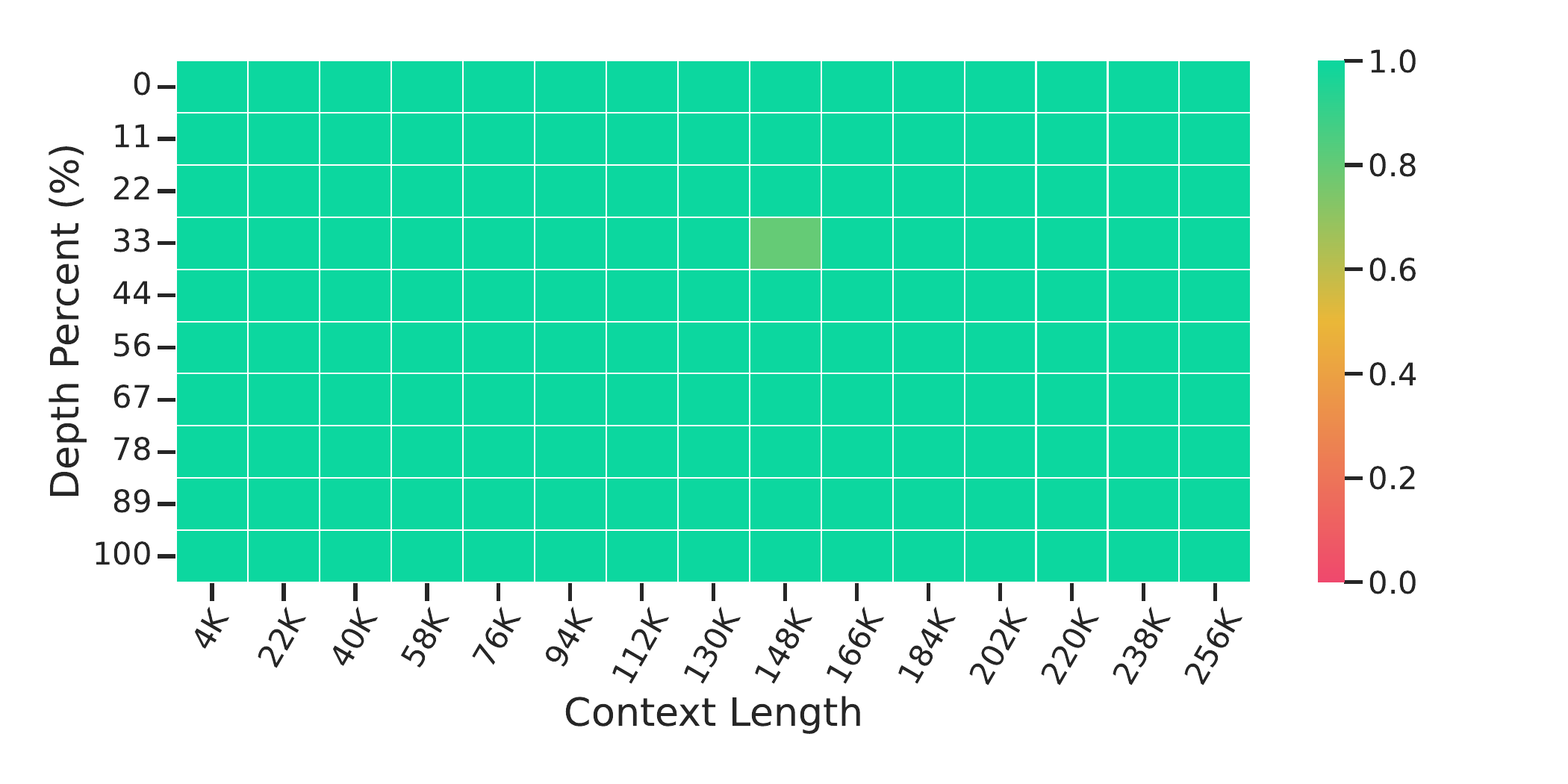}
        \caption{SnapKV, KV budget = 2048.}
        \label{fig:niah-results-qwen257b-snapkv}
    \end{subfigure}
    \begin{subfigure}{0.33\textwidth}
        \centering
        \includegraphics[width=\linewidth]{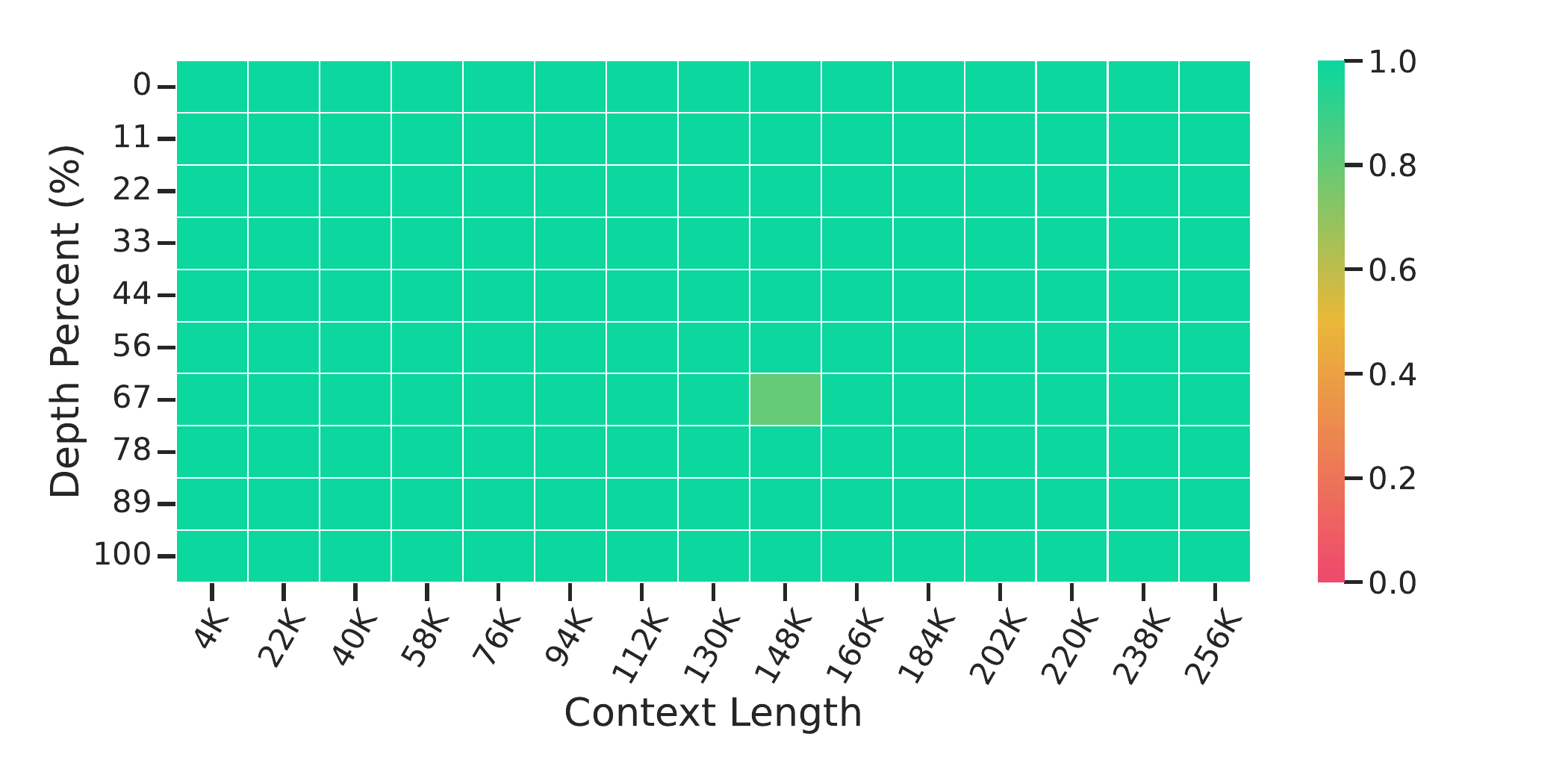}
        \caption{FastKV, KV budget = 2048.}
        \label{fig:niah-results-qwen257b-fastkv}
    \end{subfigure}
    \begin{subfigure}{0.33\textwidth}
        \centering
        \includegraphics[width=\linewidth]{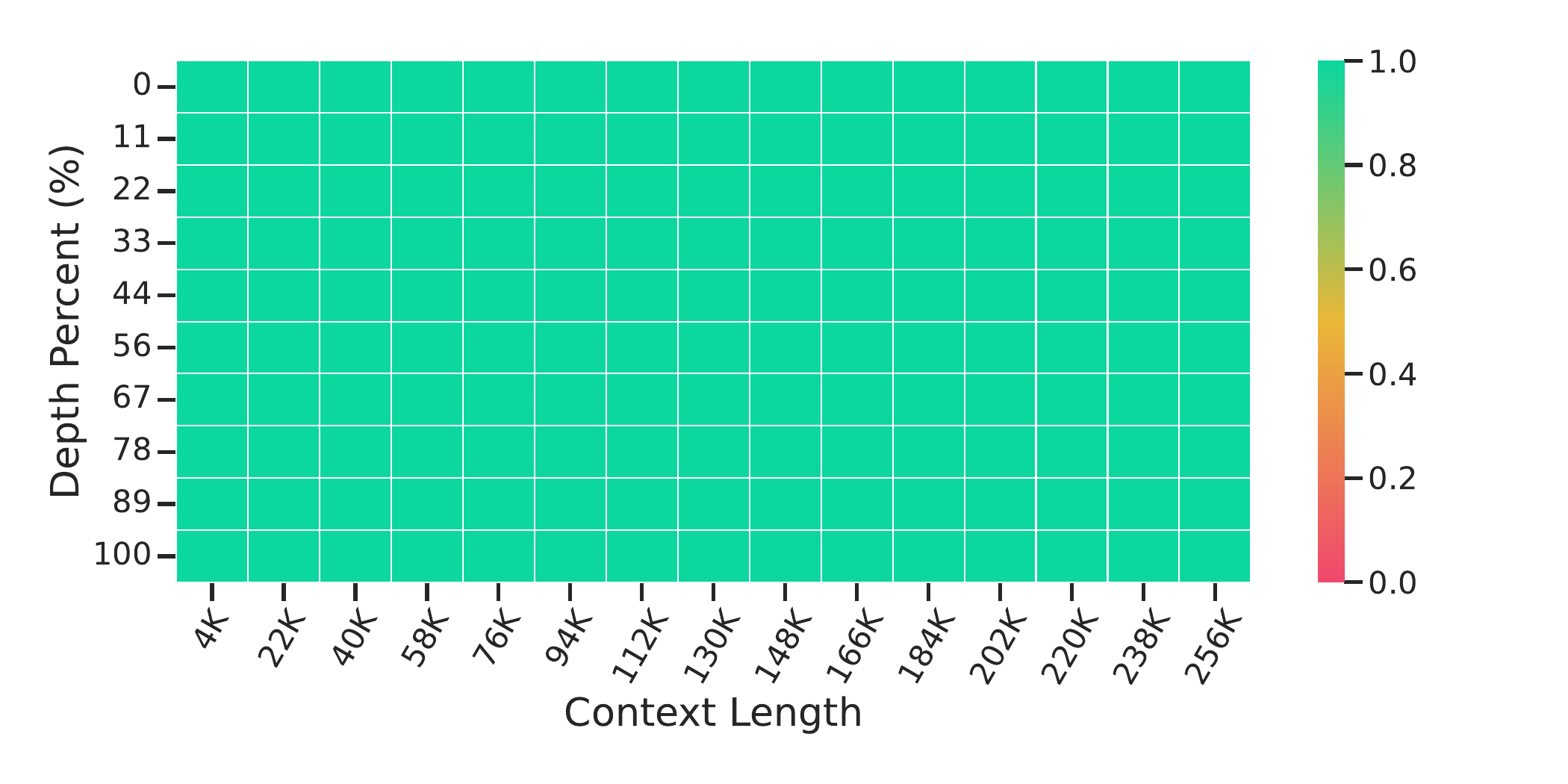}
        \caption{ASL, KV budget = 2048.}
        \label{fig:niah-results-qwen257b-asl}
    \end{subfigure}
    \begin{subfigure}{0.33\textwidth}
        \centering
        \includegraphics[width=\linewidth]{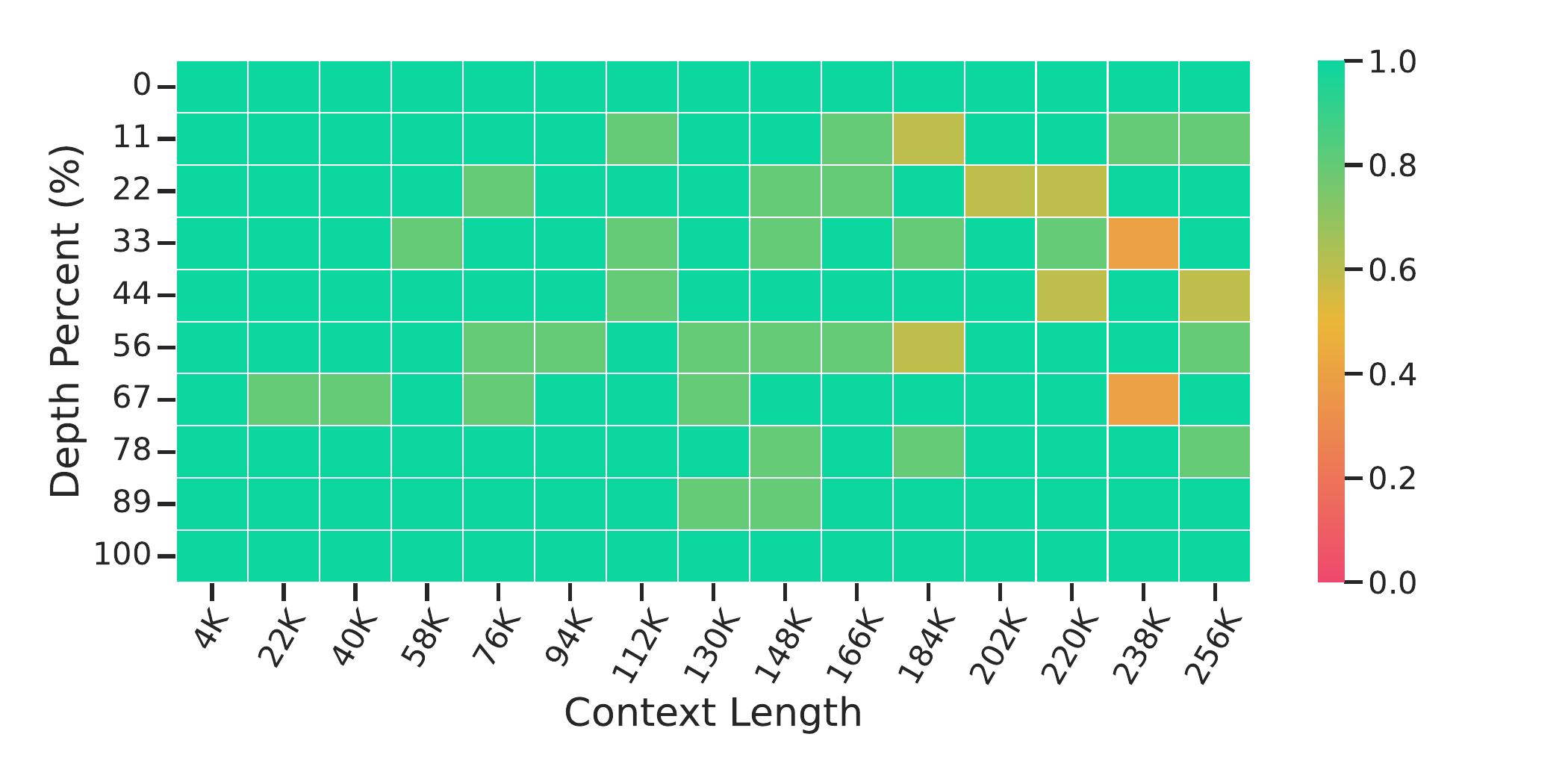}
        \caption{GemFilter, KV budget = 2048.}
        \label{fig:niah-results-qwen257b-gemfilter}
    \end{subfigure}
    \begin{subfigure}{0.33\textwidth}
        \centering
        \includegraphics[width=\linewidth]{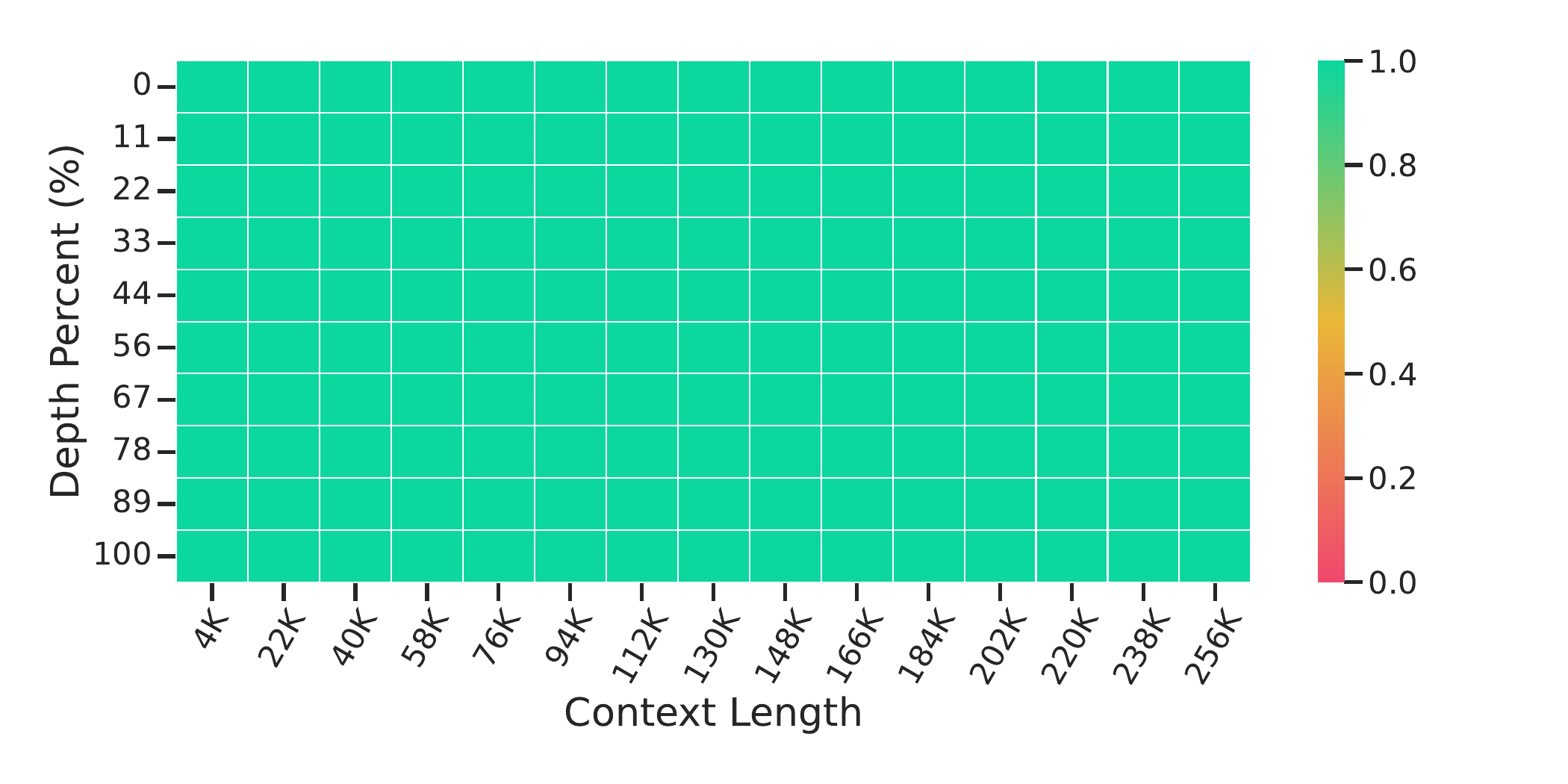}
        \caption{\aslgf, KV budget = 2048.}
        \label{fig:niah-results-qwen257b-asl2pass}
    \end{subfigure}
    \caption{NIAH results of Qwen2.5-7B.}
    \label{fig:niah-results-qwen257b}
\end{figure*}

\begin{table}[t]
    \small
    \centering
    \setlength{\tabcolsep}{1ex}
    \caption{TTFT ($\downarrow$) comparison on RULER, averaged over 13 tasks. Ratio to Full KV ($= 1$) is reported.}
    \resizebox{\linewidth}{!}{
    \begin{tabular}{l|cccccc}
    \toprule
    \textbf{Methods}       & \textbf{4k} & \textbf{8k} & \textbf{16k} & \textbf{32k} & \textbf{64k} & \textbf{128k} \\
    \midrule
    \midrule
    \multicolumn{7}{c}{\textbf{Llama-3.1-8B-UL, KV Budget = 2048.}} \\
    \midrule
    SnapKV        & 1.08 & 1.05 & 1.04 & 1.03 & 1.01 & 1.01 \\
    FastKV        & \textbf{0.85} & \textbf{0.66} & 0.58 & 0.53 & 0.51 & 0.50 \\
    PyramidInfer  & 2.09 & 3.38 & OOM  & OOM  & OOM  & OOM  \\
    \rowcolor[HTML]{DAE8FC}
    ASL           & 1.06 & 0.92 & 0.82 & 0.61 & 0.67 & 0.79 \\
    GemFilter     & 1.11 & 0.74 & \textbf{0.57} & \textbf{0.49} & \textbf{0.45} & \textbf{0.44} \\
    \rowcolor[HTML]{DAE8FC}
    \aslgf        & 1.65  & 1.17  & 0.92  & 0.65  & 0.69  & 0.80 \\
    \midrule
    \midrule
    \multicolumn{7}{c}{\textbf{Qwen2.5-7B, KV Budget = 2048.}} \\
    \midrule
    SnapKV        & 1.07 & 1.04 & 1.03 & 1.02 & 1.01 & 1.01 \\
    FastKV        & \textbf{0.87} & \textbf{0.69} & \textbf{0.61} & \textbf{0.56} & \textbf{0.55} & 0.54 \\
    PyramidInfer  & 1.94 & 3.15 & OOM  & OOM  & OOM  & OOM  \\
    \rowcolor[HTML]{DAE8FC}
    ASL           & 1.12 & 1.03 & 0.93 & 0.83 & 0.79 & 0.81 \\
    GemFilter     & 1.19 & 0.83 & 0.66 & 0.57 & \textbf{0.55} & \textbf{0.53} \\
    \rowcolor[HTML]{DAE8FC}
    \aslgf        & 1.73 & 1.43 & 1.25 & 1.48 & 0.79 & 0.95 \\
    \bottomrule
    \end{tabular}
    }
    \label{tab:ttft-ruler}
\end{table}

\begin{table}[t]
    \small
    \centering
    \setlength{\tabcolsep}{1ex}
    \caption{TPOT ($\downarrow$) comparison on RULER, averaged over 13 tasks. Ratio to Full KV ($= 1$) is reported.}
    \resizebox{\linewidth}{!}{
    \begin{tabular}{l|cccccc}
    \toprule
    \textbf{Methods}       & \textbf{4k} & \textbf{8k} & \textbf{16k} & \textbf{32k} & \textbf{64k} & \textbf{128k} \\
    \midrule
    \midrule
    \multicolumn{7}{c}{\textbf{Llama-3.1-8B-UL, KV Budget = 2048.}} \\
    \midrule
    SnapKV        & 0.98 & 0.94 & 0.83 & 0.68 & 0.48 & 0.30 \\
    FastKV        & 0.91 & 0.85 & 0.77 & 0.62 & 0.43 & 0.27 \\
    PyramidInfer  & 1.57 & 1.63 & OOM  & OOM  & OOM  & OOM  \\
    \rowcolor[HTML]{DAE8FC}
    ASL           & 0.89 & 0.86 & 0.77 & 0.62 & 0.44 & 0.28 \\
    GemFilter     & 0.81 & 0.80 & 0.71 & 0.58 & \textbf{0.40} & \textbf{0.25} \\
    \rowcolor[HTML]{DAE8FC}
    \aslgf        & \textbf{0.80} & \textbf{0.77} & \textbf{0.69} & \textbf{0.57}  & 0.41  & \textbf{0.25} \\
    \midrule
    \midrule
    \multicolumn{7}{c}{\textbf{Qwen2.5-7B, KV Budget = 2048.}} \\
    \midrule
    SnapKV        & 0.98 & 0.87 & 0.70 & 0.48 & 0.29 & 0.15 \\
    FastKV        & 0.99 & 0.88 & 0.71 & 0.48 & 0.30 & 0.15 \\
    PyramidInfer  & 1.39 & 1.32 & OOM  & OOM  & OOM  & OOM  \\
    \rowcolor[HTML]{DAE8FC}
    ASL           & 1.01 & 0.89 & 0.72 & 0.49 & 0.30 & 0.15 \\
    GemFilter     & \textbf{0.76} & \textbf{0.68} & \textbf{0.55} & \textbf{0.37} & \textbf{0.23} & \textbf{0.11} \\
    \rowcolor[HTML]{DAE8FC}
    \aslgf        & \textbf{0.76} & 0.72 & 0.62 & 0.55 & \textbf{0.23} & 0.13 \\
    \bottomrule
    \end{tabular}
    }
    \label{tab:tpot-ruler}
\end{table}
\begin{table}[t] 
  \small 
  \centering 
  \setlength{\tabcolsep}{1ex} 
  \caption{Throughput ($\uparrow$).} \resizebox{0.4\textwidth}{!}{ \begin{tabular}{l|cccc} 
  \toprule 
  \textbf{Methods} & \textbf{Mean} & \textbf{Median} & \textbf{95th} & \textbf{99th}  \\
    \midrule
    \midrule
    \multicolumn{5}{c}{\textbf{Llama-3.1-8B-UL, KV Budget = 2048.}} \\
    \midrule
    FastKV        & 3.07 & 1.84 & 0.49 & 0.36  \\
    \rowcolor[HTML]{DAE8FC}
    ASL           & 2.27 & 0.98 & 0.29 & 0.26  \\
    GemFilter     & 2.36 & 1.60 & 0.28 & 0.14  \\
    \rowcolor[HTML]{DAE8FC}
    \aslgf        & 1.57 &0.70 &0.18 &0.09  \\
    \midrule
    \midrule
    \multicolumn{5}{c}{\textbf{Qwen2.5-7B, KV Budget = 2048.}} \\
    \midrule
    FastKV        & 3.43 & 2.19 & 0.57 & 0.29  \\
    \rowcolor[HTML]{DAE8FC}
    ASL           & 2.35 & 1.54 & 0.40 & 0.21  \\
    GemFilter     & 3.32 & 2.20 & 0.41 & 0.14  \\
    \rowcolor[HTML]{DAE8FC}
    \aslgf        & 1.82 & 1.82 & 0.29  & 0.11  \\
    \bottomrule
  \end{tabular}
  }
  \label{tab:throughput}
\end{table}

\begin{table}[t]
    \small
    \centering
    \setlength{\tabcolsep}{0.5ex}
    \caption{Memory usage ($\downarrow$) comparison, averaged over 13 tasks, numbers reported in GB.}
    \resizebox{\linewidth}{!}{
    \begin{tabular}{l|cccccc}
    \toprule
    \textbf{Models} & \textbf{Full KV} & \textbf{SnapKV} & \textbf{FastKV} & \textbf{ASL} & \textbf{GemFilter} & \textbf{\aslgf} \\
    \midrule
    \midrule
    \multicolumn{7}{c}{\textbf{InfiniteBench, average context length = 214k, KV budget = 2048.}} \\
    \midrule
    Llama-3.1-8B-UL & 18.6 & \textbf{0.3} & \textbf{0.3} & {\cellcolor[HTML]{DAE8FC}}\textbf{0.3} & \textbf{0.3} & {\cellcolor[HTML]{DAE8FC}}\textbf{0.3} \\
    Qwen2.5-7B      & 8.6 & \textbf{0.2} & \textbf{0.2} & {\cellcolor[HTML]{DAE8FC}}\textbf{0.2} & \textbf{0.2} & {\cellcolor[HTML]{DAE8FC}}\textbf{0.2} \\
    \midrule
    \midrule
    \multicolumn{7}{c}{\textbf{RULER, context length = 128k, KV budget = 2048.}} \\
    \midrule
    Llama-3.1-8B-UL & 17.1 & \textbf{0.3} & \textbf{0.3} & {\cellcolor[HTML]{DAE8FC}}\textbf{0.3} & \textbf{0.3} & {\cellcolor[HTML]{DAE8FC}}\textbf{0.3} \\
    Qwen2.5-7B      & 7.5 & \textbf{0.2} & \textbf{0.2} & {\cellcolor[HTML]{DAE8FC}}\textbf{0.2} & \textbf{0.2} & {\cellcolor[HTML]{DAE8FC}}\textbf{0.2} \\
    \bottomrule
    \end{tabular}
    }
    \label{tab:memory}
\end{table}
\begin{table}[t]
    \small
    \centering
    \setlength{\tabcolsep}{1ex}
    \caption{Performance of varying KV budgets, Qwen2.5-7B, RULER, 128k. 
    TTFT and TPOT are reported in ratio to Full KV ($= 1$). Memory usage is reported in GB.}
    \resizebox{\linewidth}{!}{
    \begin{tabular}{l|cccc}
    \toprule
    \textbf{Methods} & \textbf{Accuracy ($\uparrow$)} & \textbf{TTFT ($\downarrow$)} & \textbf{TPOT ($\downarrow$)} & \textbf{Memory ($\downarrow$)} \\
    \midrule
    \multicolumn{5}{c}{\textbf{KV Budget = 2048.}} \\
    \midrule
    FastKV        & 59.1 & 0.54 & 0.15 & \textbf{0.2} \\
    \rowcolor[HTML]{DAE8FC}
    ASL           & \textbf{66.4} & 0.81 & 0.15 & \textbf{0.2} \\
    GemFilter     & 56.7 & \textbf{0.53} & \textbf{0.11} & \textbf{0.2} \\
    \rowcolor[HTML]{DAE8FC}
    \aslgf        & 62.5 & 0.95 & 0.13 & \textbf{0.2} \\
    \midrule
    \multicolumn{5}{c}{\textbf{KV Budget = 8192.}} \\
    \midrule
    FastKV        & 67.0 & \textbf{0.55} & 0.18 & \textbf{0.5} \\
    \rowcolor[HTML]{DAE8FC}
    ASL           & \textbf{68.5} & 0.94 & 0.18 & 0.6 \\
    GemFilter     & 59.9 & \textbf{0.55} & \textbf{0.11} & \textbf{0.5} \\
    \rowcolor[HTML]{DAE8FC}
    \aslgf        & 66.5 & 0.96 & \textbf{0.11} & \textbf{0.5} \\
    \bottomrule
    \end{tabular}
    }
    \label{tab:budget}
\end{table}
\begin{table}[t]
    \small
    \centering
    \setlength{\tabcolsep}{1.5ex}
    \caption{Effect of relative variance threshold $\tau$, RULER, 128k. 
    TTFT is reported in ratio to Full KV ($= 1$).}
    \resizebox{\linewidth}{!}{
    \begin{tabular}{l|ccccc}
    \toprule
    \textbf{Metrics} & $\tau = 0.2$ & $\tau = 0.3$ & $\tau = 0.4$ & $\tau = 0.5$ & $\tau = 0.6$ \\
    \midrule
    \midrule
    \multicolumn{6}{c}{\textbf{Llama-3.1-8B-UL, KV budget = 2048.}} \\
    \midrule
    Accuracy ($\uparrow$) & 55.4 & 56.1 & 55.1 & 57.1 & \textbf{57.9} \\
    TTFT ($\downarrow$)    & 0.93 & 0.79 & 0.66 & 0.60 & \textbf{0.57} \\
    \midrule
    \multicolumn{6}{c}{\textbf{Llama-3.1-8B-UL, KV Budget = Full (before sel.) \& 2048 (after sel.)}} \\
    \midrule
    Accuracy ($\uparrow$)  & 68.6 & \textbf{69.2} & 63.2 & 63.6 & 63.8 \\
    TTFT ($\downarrow$)    & 0.92 & 0.78 & 0.66 & 0.61 & \textbf{0.57} \\
    \midrule
    \multicolumn{6}{c}{\textbf{Qwen2.5-7B, KV budget = 2048.}} \\
    \midrule
    Accuracy ($\uparrow$)  & 64.9 & 66.5 & \textbf{66.9} & 65.8 & 62.1 \\
    TTFT ($\downarrow$)    & 0.90 & 0.83 & 0.78 & 0.75 & \textbf{0.73} \\
    \midrule
    \multicolumn{6}{c}{\textbf{Qwen2.5-7B, KV Budget = Full (before sel.) \& 2048 (after sel.)}} \\
    \midrule
    Accuracy ($\uparrow$)  & \textbf{78.2} & 74.2 & 70.5 & 67.5 & 63.3 \\
    TTFT ($\downarrow$)    & 0.92 & 0.81 & 0.76 & 0.71 & \textbf{0.67} \\
    \bottomrule
    \end{tabular}
    }
    \label{tab:ablt-th}
\end{table}

\subsection{Accuracy Evaluation}
\myparagraph{InfiniteBench}
Table~\ref{tab:score-infinitebench} shows the accuracy on InfiniteBench across 10 tasks. 
Under the KV budget of 2048, \aslgf achieves the highest average score for Llama-3.1-8B-UL. For Qwen2.5-7B, ASL reports the highest average score, tying FastKV. 
When using full KV before token selection, the gap between adaptive selection (ASL) and user-specified selection (FastKV) is more substantial, especially for hard tasks such as KV retrieval, where the performance of ASL is close to full KV while FastKV is much inferior. 

\myparagraph{RULER}
Table~\ref{tab:score-ruler} reports the accuracy on RULER with various context lengths.
For Llama-3.1-8B-UL, ASL consistently outperforms FastKV. However, \aslgf generally performs worse than GemFilter, suggesting less compatibility of ASL with this ultra long variant of Llama 3.1 by NVidia. 
For Qwen2.5-7B, the advantage of ASL over FastKV is observed when the context length $\geq$ 16k, and 
\aslgf performs better than GemFilter for over 8k contexts. 
Moreover, the gaps between ASL and existing methods are more substantial for longer contexts such as 128k. 
Like InfiniteBench, ASL's superiority over FastKV is also more remarkable when using full KV before token selection. 

\myparagraph{NIAH}
Figure~\ref{fig:niah-results-qwen257b} shows the results across various context lengths on NIAH, using Qwen2.5-7B. ASL and \aslgf report full scores for all the context lengths, delivering the same performance as full KV. SnapKV and FastKV retrieve almost all needles except at the context length of 148k. In contrast, GemFilter reports mediocre performance, especially for long contexts. 

\subsection{Efficiency Evaluation}
\myparagraph{TTFT}
Table~\ref{tab:ttft-ruler} reports the time to first token (TTFT), varying the context length on RULER. To show how the time scales with the context length, we report the numbers by calculating the ratio to full KV's TTFT. 
SnapKV, which does not optimize TTFT, report almost the same result as full KV. 
FastKV, GemFilter, and ASL methods are faster than SnapKV, with GemFilter being the fastest due to its smallest value of selection layer. ASL is slower than FastKV because its token selection is usually at deeper layers than FastKV's. 

\myparagraph{TPOT}
Table~\ref{tab:tpot-ruler} reports the time per output token (TPOT), varying context lengths on RULER. Like TTFT, the numbers are also reported in the ratio to full KV. 
All the KV cache reduction methods exhibit faster decoding speed than full KV. 
For Llama-3.1-8B, FastKV, GemFilter and ASL methods are faster than SnapKV, and similar TPOTs are observed for these methods under longer contexts. 
For Qwen2.5-7B, SnapKV, FastKV, and ASL report similar TPOTs, whereas GemFilter becomes the fastest and \aslgf shows competitive speed for context lengths over 64k.   


\myparagraph{Throughput}
Table~\ref{tab:throughput} shows the throughputs of ASL, FastKV, and GemFilter on RULER, 128k context, 2048 KV budget. We report mean, median, 95th percentile, and 99th percentile. The latter two refer to the slowest 5\% and 1\% queries, respectively. 
ASL trails FastKV and GemFilter, with mean throughputs of 74\% (Llama-3.1-8B-UL) and 69\% (Qwen2.5-7B) compared to FastKV. ASL\_2pass reports mean throughputs of 67\% (Llama-3.1-8B-UL) and 55\% (Qwen2.5-7B)  compared to GemFilter. For the slowest queries on Qwen2.5-7B, the gap between ASL and FastKV/GemFilter becomes smaller, e.g., 72\% throughputs compared to FastKV and 79\% compared to GemFilter.

ASL trades throughput for accuracy, as shown in Table~\ref{tab:budget} for the above setting (Qwen-2.5-7B), where substantial accuracy improvement can be seen. 
In addition, the number of output tokens for the RULER benchmark is small (27 for mean and 120 for 99th percentile). For real-world applications, in which the number of output tokens is often much larger, the gap between the throughputs of ASL and FastKV/GemFilter tends to be even smaller, because most processing time will be spent on the decoding stage and these methods report almost the same TPOT (Table~\ref{tab:tpot-ruler}).

\myparagraph{Memory Usage}
Table~\ref{tab:memory} reports the memory usage. 
All the KV cache reduction methods consume far less memory than full KV, and their memory usages are almost identical. 
This showcases that ASL's additional overhead in memory is negligible. 

\subsection{Varying KV Budget}
Table~\ref{tab:budget} shows the performance under two KV budget settings: 2048 and 8192. 
By trading TTFT, ASL and \aslgf consistently exhibit higher accuracy than their counterparts FastKV and GemFilter for the two budget settings. Moreover, ASL methods report competitive TPOTs and consume around the same size of memory as others. 

\subsection{Effect of Relative Variance Threshold}
We vary the relative variance threshold $\tau$ from 0.2 to 0.6, and report the results on RULER, 128k in Table~\ref{tab:ablt-th}. 
The accuracy for different tasks is reported in Appendix~\ref{sec:exp-effect-tau-tasks}. 
TPOT and memory usage are barely affected by $\tau$ and not reported. 
The accuracy fluctuates for Llama-3.1-8B-UL when $\tau$ varies from 0.2 to 0.6. For Qwen2.5-7B, under the KV budget of 2048, the accuracy increases and then decreases, and when full KV is available before token selection, a generally decreasing trend of accuracy is observed. 
TTFT consistently decreases under larger thresholds, because relative variance decreases as layers, meeting a larger threshold first. 
Seeing the results, we suggest using $\tau = 0.3$ for the trade-off between accuracy and TTFT. 

\section{Conclusion}
\label{sec:concl}
We proposed ASL, a KV cache reduction method that adaptively chooses the selection layer for layer-wise token pruning in various tasks. 
Observing attention patterns across tasks, we exploited the variance of token ranks ordered by attention score. 
ASL works in a one-shot selection manner, selecting tokens at a layer and propagating only those tokens to deeper layers. 
To meet the KV budget requirement, we jointly used ASL with existing KV cache reduction methods SnapKV and GemFilter. 
We evaluated ASL on three benchmarks. 
The results demonstrated its accuracy-efficiency trade-offs compared to state-of-the-art layer-wise token pruning methods.


\section*{Limitations}
In this work, we focus on applying ASL to one-shot methods such as FastKV and GemFilter. There are other layer-wise token pruning methods, such as LasyLLM and OmniKV, as summarized in Table~\ref{tab:layer-wise-methods}. These methods select tokens multiple times, and the way of integrating ASL into these methods is yet to be investigated. One possible solution is to use multiple variance thresholds, or a threshold with a decay factor. 

There are also numerous methods for KV cache reduction that do not belong to the category of layer-wise token pruning, as noted in Section~\ref{sec:prelim}. A more comprehensive comparison with those methods may better reveal the positioning of this work. 

Another limitation is that only two LLMs, Llama 3.1-8B-UL and Qwen2.5-7B, have been evaluated for the proposed method. Less competitiveness of ASL, when integrated with GemFilter on Llama 3.1-8B-UL, has been observed. 
\section*{Acknowledgements}
This work is supported by JSPS Kakenhi JP23K17456, JP23K25157, JP23K28096, and JP25H01117. 

\bibliography{refs}

@article{vaswani2017attention,
  title={Attention is all you need},
  author={Vaswani, Ashish and Shazeer, Noam and Parmar, Niki and Uszkoreit, Jakob and Jones, Llion and Gomez, Aidan N and Kaiser, {\L}ukasz and Polosukhin, Illia},
  journal={Advances in neural information processing systems},
  volume={30},
  year={2017}
}

@misc{awesome,
  author = {Longze Chen},
  title = {Awesome-KV-Cache-Compression},
  howpublished = "\url{https://github.com/October2001/Awesome-KV-Cache-Compression}",
  year = {2024}
}

@article{tang2024quest,
  title={Quest: Query-aware sparsity for efficient long-context {LLM} inference},
  author={Tang, Jiaming and Zhao, Yilong and Zhu, Kan and Xiao, Guangxuan and Kasikci, Baris and Han, Song},
  journal={arXiv preprint arXiv:2406.10774},
  year={2024}
}

@article{singhania2024loki,
  title={Loki: Low-rank keys for efficient sparse attention},
  author={Singhania, Prajwal and Singh, Siddharth and He, Shwai and Feizi, Soheil and Bhatele, Abhinav},
  journal={Advances in Neural Information Processing Systems},
  volume={37},
  pages={16692--16723},
  year={2024}
}

@article{chen2024magicpig,
  title={{MagicPig}: {LSH} sampling for efficient {LLM} generation},
  author={Chen, Zhuoming and Sadhukhan, Ranajoy and Ye, Zihao and Zhou, Yang and Zhang, Jianyu and Nolte, Niklas and Tian, Yuandong and Douze, Matthijs and Bottou, Leon and Jia, Zhihao and others},
  journal={arXiv preprint arXiv:2410.16179},
  year={2024}
}

@article{jiang2024minference,
  title={{MInference} 1.0: Accelerating pre-filling for long-context {LLMs} via dynamic sparse attention},
  author={Huiqiang Jiang and Yucheng Li and Chengruidong Zhang and Qianhui Wu and Xufang Luo and Surin Ahn and Zhenhua Han and Amir H. Abdi and Dongsheng Li and Chin-Yew Lin and Yuqing Yang and Lili Qiu},
  journal={Advances in Neural Information Processing Systems},
  volume={37},
  pages={52481--52515},
  year={2024}
}

@misc{minference,
  author = {Huiqiang Jiang and Yucheng Li and Chengruidong Zhang and Qianhui Wu and Xufang Luo and Surin Ahn and Zhenhua Han and Amir H. Abdi and Dongsheng Li and Chin-Yew Lin and Yuqing Yang and Lili Qiu},
  title = {MInference},
  howpublished = "\url{https://github.com/microsoft/MInference/tree/main}",
  year = {2024}
}

@article{xiao2023efficient,
  title={Efficient streaming language models with attention sinks},
  author={Xiao, Guangxuan and Tian, Yuandong and Chen, Beidi and Han, Song and Lewis, Mike},
  journal={arXiv preprint arXiv:2309.17453},
  year={2023}
}

@article{zhang2023h2o,
  title={{H2O}: Heavy-hitter oracle for efficient generative inference of large language models},
  author={Zhang, Zhenyu and Sheng, Ying and Zhou, Tianyi and Chen, Tianlong and Zheng, Lianmin and Cai, Ruisi and Song, Zhao and Tian, Yuandong and R{\'e}, Christopher and Barrett, Clark and others},
  journal={Advances in Neural Information Processing Systems},
  volume={36},
  pages={34661--34710},
  year={2023}
}

@article{ge2023model,
  title={Model tells you what to discard: Adaptive {KV} cache compression for {LLMs}},
  author={Ge, Suyu and Zhang, Yunan and Liu, Liyuan and Zhang, Minjia and Han, Jiawei and Gao, Jianfeng},
  journal={arXiv preprint arXiv:2310.01801},
  year={2023}
}

@article{li2024snapkv,
  title={{SnapKV}: {LLM} knows what you are looking for before generation},
  author={Li, Yuhong and Huang, Yingbing and Yang, Bowen and Venkitesh, Bharat and Locatelli, Acyr and Ye, Hanchen and Cai, Tianle and Lewis, Patrick and Chen, Deming},
  journal={Advances in Neural Information Processing Systems},
  volume={37},
  pages={22947--22970},
  year={2024}
}

@article{feng2024ada,
  title={{Ada-KV}: Optimizing {KV} cache eviction by adaptive budget allocation for efficient {LLM} inference},
  author={Feng, Yuan and Lv, Junlin and Cao, Yukun and Xie, Xike and Zhou, S Kevin},
  journal={arXiv preprint arXiv:2407.11550},
  year={2024}
}

@article{fu2024not,
  title={Not all heads matter: A head-level {KV} cache compression method with integrated retrieval and reasoning},
  author={Fu, Yu and Cai, Zefan and Asi, Abedelkadir and Xiong, Wayne and Dong, Yue and Xiao, Wen},
  journal={arXiv preprint arXiv:2410.19258},
  year={2024}
}

@article{gromov2024unreasonable,
  title={The unreasonable ineffectiveness of the deeper layers},
  author={Gromov, Andrey and Tirumala, Kushal and Shapourian, Hassan and Glorioso, Paolo and Roberts, Daniel A},
  journal={arXiv preprint arXiv:2403.17887},
  year={2024}
}

@article{liu2024minicache,
  title={{MiniCache}: {KV} cache compression in depth dimension for large language models},
  author={Liu, Akide and Liu, Jing and Pan, Zizheng and He, Yefei and Haffari, Reza and Zhuang, Bohan},
  journal={Advances in Neural Information Processing Systems},
  volume={37},
  pages={139997--140031},
  year={2024}
}

@article{liu2024foldgpt,
  title={{FoldGPT}: Simple and effective large language model compression scheme},
  author={Liu, Songwei and Zeng, Chao and Li, Lianqiang and Yan, Chenqian and Fu, Lean and Mei, Xing and Chen, Fangmin},
  journal={arXiv preprint arXiv:2407.00928},
  year={2024}
}

@article{qiao2024swiftkv,
  title={{SwiftKV}: Fast Prefill-Optimized Inference with Knowledge-Preserving Model Transformation},
  author={Qiao, Aurick and Yao, Zhewei and Rajbhandari, Samyam and He, Yuxiong},
  journal={arXiv preprint arXiv:2410.03960},
  year={2024}
}

@article{wang2024squeezeattention,
  title={{SqueezeAttention}: 2d management of {KV}-cache in {LLM} inference via layer-wise optimal budget},
  author={Wang, Zihao and Cui, Bin and Gan, Shaoduo},
  journal={arXiv preprint arXiv:2404.04793},
  year={2024}
}

@article{yang2024pyramidinfer,
  title={{PyramidInfer}: Pyramid {KV} cache compression for high-throughput {LLM} inference},
  author={Yang, Dongjie and Han, XiaoDong and Gao, Yan and Hu, Yao and Zhang, Shilin and Zhao, Hai},
  journal={arXiv preprint arXiv:2405.12532},
  year={2024}
}

@article{cai2024pyramidkv,
  title={{PyramidKV}: Dynamic {KV} cache compression based on pyramidal information funneling},
  author={Cai, Zefan and Zhang, Yichi and Gao, Bofei and Liu, Yuliang and Liu, Tianyu and Lu, Keming and Xiong, Wayne and Dong, Yue and Chang, Baobao and Hu, Junjie and others},
  journal={arXiv preprint arXiv:2406.02069},
  year={2024}
}

@article{shi2024discovering,
  title={Discovering the gems in early layers: Accelerating long-context {LLMs} with 1000x input token reduction},
  author={Shi, Zhenmei and Ming, Yifei and Nguyen, Xuan-Phi and Liang, Yingyu and Joty, Shafiq},
  journal={arXiv preprint arXiv:2409.17422},
  year={2024}
}

@article{zhou2024dynamickv,
  title={{DynamicKV}: Task-Aware Adaptive {KV} Cache Compression for Long Context {LLMs}},
  author={Zhou, Xiabin and Wang, Wenbin and Zeng, Minyan and Guo, Jiaxian and Liu, Xuebo and Shen, Li and Zhang, Min and Ding, Liang},
  journal={arXiv preprint arXiv:2412.14838},
  year={2024}
}

@article{fu2024lazyllm,
  title={{LazyLLM}: Dynamic token pruning for efficient long context {LLM} inference},
  author={Fu, Qichen and Cho, Minsik and Merth, Thomas and Mehta, Sachin and Rastegari, Mohammad and Najibi, Mahyar},
  journal={arXiv preprint arXiv:2407.14057},
  year={2024}
}

@inproceedings{hao2025omnikv,
  title={{OmniKV}: Dynamic context selection for efficient long-context {LLMs}},
  author={Hao, Jitai and Zhu, Yuke and Wang, Tian and Yu, Jun and Xin, Xin and Zheng, Bo and Ren, Zhaochun and Guo, Sheng},
  booktitle={The Thirteenth International Conference on Learning Representations},
  year={2025}
}

@article{jo2025fastkv,
  title={{FastKV}: {KV} Cache Compression for Fast Long-Context Processing with Token-Selective Propagation},
  author={Jo, Dongwon and Song, Jiwon and Kim, Yulhwa and Kim, Jae-Joon},
  journal={arXiv preprint arXiv:2502.01068},
  year={2025}
}

@article{qin2025cake,
  title={{CAKE}: Cascading and adaptive {KV} cache eviction with layer preferences},
  author={Qin, Ziran and Cao, Yuchen and Lin, Mingbao and Hu, Wen and Fan, Shixuan and Cheng, Ke and Lin, Weiyao and Li, Jianguo},
  journal={arXiv preprint arXiv:2503.12491},
  year={2025}
}

@article{jin2025promptdistill,
  title={{PromptDistill}: Query-based Selective Token Retention in Intermediate Layers for Efficient Large Language Model Inference},
  author={Jin, Weisheng and Song, Maojia and Pala, Tej Deep and Chia, Yew Ken and Zadeh, Amir and Li, Chuan and Poria, Soujanya},
  journal={arXiv preprint arXiv:2503.23274},
  year={2025}
}

@article{long2025sliminfer,
  title={{SlimInfer}: Accelerating Long-Context {LLM} Inference via Dynamic Token Pruning},
  author={Long, Lingkun and Yang, Rubing and Huang, Yushi and Hui, Desheng and Zhou, Ao and Yang, Jianlei},
  journal={arXiv preprint arXiv:2508.06447},
  year={2025}
}

@article{yu2025evolkv,
  title={{EvolKV}: Evolutionary {KV} Cache Compression for {LLM} Inference},
  author={Yu, Bohan and Chai, Yekun},
  journal={arXiv preprint arXiv:2509.08315},
  year={2025}
}

@inproceedings{lee2024infinigen,
  title={{InfiniGen}: Efficient generative inference of large language models with dynamic {KV} cache management},
  author={Lee, Wonbeom and Lee, Jungi and Seo, Junghwan and Sim, Jaewoong},
  booktitle={18th USENIX Symposium on Operating Systems Design and Implementation (OSDI 24)},
  pages={155--172},
  year={2024}
}

@article{sun2024shadowkv,
  title={{ShadowKV}: {KV} cache in shadows for high-throughput long-context {LLM} inference},
  author={Sun, Hanshi and Chang, Li-Wen and Bao, Wenlei and Zheng, Size and Zheng, Ningxin and Liu, Xin and Dong, Harry and Chi, Yuejie and Chen, Beidi},
  journal={arXiv preprint arXiv:2410.21465},
  year={2024}
}

@article{liu2024kivi,
  title={{KIVI}: A tuning-free asymmetric 2bit quantization for {KV} cache},
  author={Liu, Zirui and Yuan, Jiayi and Jin, Hongye and Zhong, Shaochen and Xu, Zhaozhuo and Braverman, Vladimir and Chen, Beidi and Hu, Xia},
  journal={arXiv preprint arXiv:2402.02750},
  year={2024}
}

@article{hooper2024kvquant,
  title={{KVQuant}: Towards 10 million context length {LLM} inference with {KV} cache quantization},
  author={Hooper, Coleman and Kim, Sehoon and Mohammadzadeh, Hiva and Mahoney, Michael W and Shao, Sophia and Keutzer, Kurt and Gholami, Amir},
  journal={Advances in Neural Information Processing Systems},
  volume={37},
  pages={1270--1303},
  year={2024}
}

@inproceedings{zhang2024bench,
  title={$\infty$ Bench: Extending long context evaluation beyond 100k tokens},
  author={Zhang, Xinrong and Chen, Yingfa and Hu, Shengding and Xu, Zihang and Chen, Junhao and Hao, Moo and Han, Xu and Thai, Zhen and Wang, Shuo and Liu, Zhiyuan and others},
  booktitle={Proceedings of the 62nd Annual Meeting of the Association for Computational Linguistics (Volume 1: Long Papers)},
  pages={15262--15277},
  year={2024}
}

@article{hsieh2024ruler,
  title={RULER: What's the Real Context Size of Your Long-Context Language Models?},
  author={Hsieh, Cheng-Ping and Sun, Simeng and Kriman, Samuel and Acharya, Shantanu and Rekesh, Dima and Jia, Fei and Zhang, Yang and Ginsburg, Boris},
  journal={arXiv preprint arXiv:2404.06654},
  year={2024}
}

@misc{niah,
  author = {Greg Kamradt},
  title = {Needle In A Haystack -- Pressure Testing {LLMs}},
  howpublished = "\url{https://github.com/gkamradt/LLMTest_NeedleInAHaystack}",
  year = {2024}
}

@inproceedings{wolf-etal-2020-transformers,
    title = "Transformers: State-of-the-Art Natural Language Processing",
    author = "Thomas Wolf and Lysandre Debut and Victor Sanh and Julien Chaumond and Clement Delangue and Anthony Moi and Pierric Cistac and Tim Rault and Rémi Louf and Morgan Funtowicz and Joe Davison and Sam Shleifer and Patrick von Platen and Clara Ma and Yacine Jernite and Julien Plu and Canwen Xu and Teven Le Scao and Sylvain Gugger and Mariama Drame and Quentin Lhoest and Alexander M. Rush",
    booktitle = "Proceedings of the 2020 Conference on Empirical Methods in Natural Language Processing: System Demonstrations",
    year = "2020",
    address = "Online",
    pages = "38--45"
}

@article{dao2023flashattention,
  title={Flashattention-2: Faster attention with better parallelism and work partitioning},
  author={Dao, Tri},
  journal={arXiv preprint arXiv:2307.08691},
  year={2023}
}

@misc{hfattention,
  title={Transformers -- Attention Interface},
  author={Huggging Face},
  howpublished = "\url{https://huggingface.co/docs/transformers/en/attention_interface}",
  year = {2025}
}

@inproceedings{charikar2002similarity,
  title={Similarity estimation techniques from rounding algorithms},
  author={Charikar, Moses S},
  booktitle={Proceedings of the thiry-fourth annual ACM symposium on Theory of computing},
  pages={380--388},
  year={2002}
}

\clearpage

\onecolumn
\appendix
\section*{Appendix}
\label{sec:appendix}

\section{Algorithm Pseudo-Codes}
\label{sec:pseudo}



Algorithm~\ref{alg:update-kv-rank} calculates pooled attention scores over the most recent tokens (specified by a window) and uses them to identify the \topk indices for each GQA head group. 
In parallel, it computes the score ranks of the pooled attention scores and stores them in a rank cache.
Compressed KV entries are returned, in line with SnapKV~\cite{li2024snapkv}. 

Based on the rank cache, Algorithm~\ref{alg:select-selection-layer} obtains the unique \topk tokens across the most recent $L_{\obs}$ layers, and then computes the relative variance of their ranks within these $L_{\obs}$ layers.
If the relative variance falls below a threshold $\tau$, the current layer is determined as the selection layer, and only the selected tokens are propagated to subsequent layers.
The rank cache is cleared afterwards.

\begin{algorithm}[h]
  \caption{GetRanks}
  \label{alg:update-kv-rank}
  \KwIn{$Q, K, V,  layer\_idx, W, S, KV\_budget, rank\_cache$} \tcp{$W$: window\_size, $S$: kernel\_size}
  \KwOut{compressed KV, update $rank\_cache[layer\_idx]$}
  $N_q \gets Len(Q)$, $B \gets BatchSize$, $H_{q, k} \gets HeadSize$, $G \gets H_q / H_k$; \tcp{GQA groups}
  $I_w \gets \{N_q - W, \dots, N_q - 1\}$; \tcp{window token indices} 
  $A \gets \mathrm{Softmax} \left(\frac{Q_{:, :, N_q - W:N_q}K^\top}{\sqrt{d}} + \mathrm{CausalMaskOnLastBlock}(W) \right)$; \tcp{$A \in \mathbb{R}^{B \times H_q \times W \times N_q}$}
  $U \gets \sum_{t = 1}^{W} A_{:, :, t, 0:N_q - W}$; \tcp{past-only histogram, $U \in \mathbb{R}^{B \times H_q \times (N_q - W)}$}
  $V \gets \mathrm{AvgPool1D}(U, \text{kernel} = S, \text{stride} = 1, \text{pad} = \lfloor S/2 \rfloor)$\;
  $P \gets \mathrm{SumGroups}\big(\mathrm{Reshape}(V, [B, H_k, G, N_q - W]),\ \text{over } G \big)$; \tcp{$P \in \mathbb{R}^{B \times H_k \times (N_q-W)}$}
  $k \gets \min(KV\_budget, N_q) - W$\;
  $I \gets \mathrm{Concat}\!\left(\mathrm{TopK}(P,k),\ I_w\right)$; \tcp{per $(B,H_k)$, add window tokens}
  $rank\_cache[layer\_idx] \gets \mathrm{RankDesc}(P)$\;
  \Return{$K[I]$,$V[I]$}\;
\end{algorithm}

\begin{algorithm}[h]
  \caption{SelectTokens}
  \label{alg:select-selection-layer}
  \KwIn{$layer\_idx, KV\_budget, W, L_{\min}, L_{\obs}, \tau, rank\_cache, init\_var$} \tcp{$W$: window\_size}
  \KwOut{$select\_idx$ (or \texttt{None}), updated $init\_var$}
  \If{$layer\_idx < L_{\min}$}{
    \Return{\texttt{None}}\;
  }
  $T \gets \mathrm{Len}(rank\_cache[layer\_idx])$; \tcp{sequence length}
  $R \gets \mathrm{Stack}\big(\{rank\_cache[l]\}_{l = layer\_idx - L_{\obs} + 1}^{layer\_idx}\big)$; \tcp{$R \in \mathbb{R}^{L_{\obs} \times T}$, collect and stack latest $L_{\obs}$ layer's rank from rank\_cache}
  $k \gets \min(KV\_budget - W, T)$\;
  $J \gets \mathrm{TopKIndices}(R, k, \text{smallest})$; \tcp{per row, $J \in \mathbb{N}^{L_{\obs} \times k}$}
  $U \gets \mathrm{Unique}(\mathrm{Flatten}(J))$\;
  $(var, init\_var) \gets \mathrm{RelNormVar}(R_{:,U}, init\_var)$; \tcp{ $var=\frac{\mathbb{E}[\mathrm{Var}_\text{layer}(R_{:,U})]}{init\_var}$,
$init\_var$ is initialized as $\mathrm{Var}_\text{layer}(R_{:,U})$ if $init\_var$=\texttt{None}
}
  \If{$var < \tau$}{
    $I_w \gets \{T, \dots, T + W - 1\}$; \tcp{ensure window tokens}
    $select\_idx \gets \mathrm{Unique}(J_{last} \cup I_w)$\;
    $rank\_cache \gets \emptyset$; \tcp{release cache}
    \Return{$select\_idx$}\;
  }
  \Else{
    \Return{\texttt{None}}\;
  }
\end{algorithm}

\section{Theoretical Analysis}
\label{sec:theoretical-analysis}
To analyze the performance of ASL, we consider a model with $L$ layers, $h$ KV heads, head dimension $d$, and context length $n$. Next, we analyze TTFT (prefilling time), TPOT (decoding time), and memory usage.

\subsection{TTFT}

For full KV, TTFT is

$$T_{\text{full}} = L \cdot T_{\text{attn}}(n)$$

where $T_{\text{attn}}(n)$ denotes the attention processing time per layer.

For ASL, let $L_\text{select}$ denote the selection layer.

From layers 0 to $L_\text{select}$, the costs are
\begin{inparaenum}[(1)]
    \item Attention: ($L_{\text{select}} + 1) \cdot T_{\text{attn}}(n)$; 
    \item Pooling: $(L_{\text{select}} - L_{\text{min}}) \cdot O(n)$; 
    \item Rank computation: $(L_{\text{select}} - L_{\text{min}}) \cdot O(n \log n)$; 
    \item Variance calculation: $(L_{\text{select}} - L_{\text{min}}) \cdot O(L_{\text{obs}} \cdot m)$, where $m$ denotes the union size and $m \leq k \cdot L_{\text{obs}}$.
\end{inparaenum}

At layer $L_\text{select}$, the cost is \topk selection: $O(n \log k)$, or $O(n)$ with QuickSelect.

From layers $L_\text{select} + 1$ to $L$, the cost is attention with selected tokens: $(L - L_{\text{select}} - 1) \cdot T_{\text{attn}}(k)$.

Therefore, ASL's TTFT is

$$T_{\text{ASL}} = (L_{\text{select}} + 1) \cdot T_{\text{attn}}(n) + (L - L_{\text{select}} - 1) \cdot T_{\text{attn}}(k) + (L_{\text{select}} - L_{\text{min}}) \cdot O(n \log n + L_{\text{obs}} \cdot m) + O(n \log k).$$

Suppose $T_{\text{attn}}(n) = O(n^2 \cdot d + n \cdot d^2)$ for FlashAttention~\citep{dao2023flashattention}. 
Because $T_{\text{attn}}(n) \gg O(n \log n)$ for large $n$ (e.g., 128k), the third and fourth terms of $T_{\text{ASL}}$ (i.e., pooling, rank,  variance calculation, and \topk selection) are negligible.

ASL's TTFT ratio to full KV is

$$\frac{T_{\text{ASL}}}{T_{\text{full}}} \approx \frac{L_{\text{select}} + 1}{L} + \frac{L - L_{\text{select}} - 1}{L} \cdot \frac{T_{\text{attn}}(k)}{T_{\text{attn}}(n)}.$$

Suppose $k = 2048$ and $n = 131072$ (128k context). $L = 32$ for Llama-3.1-8B-UL and $28$ for Qwen2.5-7B.

Under this setting, the mean $L_{\text{select}}$ is $23.9$ for Llama-3.1-8B-UL) and $21.6$ for Qwen2.5-7B (Appendix~\ref{sec:exp-selection-layer-dist} reports the the distributions of $L_{\text{select}}$).

Therefore, $\frac{T_{\text{ASL}}}{T_{\text{full}}} \approx 0.78$ (Llama-3.1-8B-UL) and $0.81$ (Qwen2.5-7B).

As for comparison, the empirical ratio in Table~\ref{tab:ttft-ruler} is $0.79$ for Llama-3.1-8B-UL and $0.81$ for Qwen2.5-7B, showcasing that the above theoretical prediction well matches the empirical result.

As for comparison, in FastKV, $L_{\text{select}} = 15$ for Llama-3.1-8B-UL and $14$ for Qwen2.5-7B. Therefore, $\frac{T_{\text{FastKV}}}{T_{\text{full}}} \approx 0.5$ and $0.54$, respectively. Empirical values in Table~\ref{tab:ttft-ruler} are $0.5$ and $0.54$, respectively, exactly matching the predicted values.

\subsection{TPOT}

For full KV, TPOT is

$$T_{\text{full}} = L \cdot O(n \cdot d).$$

For ASL, TPOT is

$$T_{\text{ASL}} = L \cdot O(k \cdot d).$$

ASL's TPOT ratio to full KV is

$$\frac{T_{\text{ASL}}}{T_{\text{full}}} = \frac{k}{n}.$$

When $k = 2048$ and $n = 131072$ (128k context), $\frac{T_{\text{ASL}}}{T_{\text{full}}} \approx 0.016$.

FastKV shares the same predication as above.

The empirical ratio of ASL to full KV in Table~\ref{tab:tpot-ruler} is $0.028$ for Llama-3.1-8B-UL and $0.015$ for Qwen2.5-7B. For Qwen2.5-7B, it roughly matches the prediction. For Llama-3.1-8B-UL, we suspect that the discrepancy is due to its model-specific implementation (an ultra long variant by NVidia).

\subsection{Memory Usage}

For full KV, memory usage is

$$M_{\text{full}} = 2 \cdot L \cdot h \cdot d \cdot n$$

where $2$ accounts for key and value.

ASL introduces two types of memory overhead: 
\begin{inparaenum}[(1)]
    \item Pooled attention scores: $M_{\text{pool}} = L_{\text{obs}} \cdot \frac{n}{w}$, where $w$ is the pooling kernel size; 
    \item Rank cache: $M_{\text{rank}} = L_{\text{obs}} \cdot \frac{n}{w}$.
\end{inparaenum}

The total additional overhead is 

$$M_{\text{ASL-OV}} = 2 \cdot L_{\text{obs}} \cdot \frac{n}{w}.$$

ASL's KV cache size under budget $k$ is

$$M_{\text{ASL-KV}} = 2 \cdot L \cdot h \cdot d \cdot k.$$

ASL's memory usage is

$$M_{\text{ASL}} = M_{\text{ASL-KV}} + M_{\text{ASL-OV}} = 2 \cdot L \cdot h \cdot d \cdot k + 2 \cdot L_{\text{obs}} \cdot \frac{n}{w}.$$

For Llama-3.1-8B-UL, $L = 32$, $h = 8$. For Qwen2.5-7B, $L = 28$, $h = 4$. $d = 128$, $w = 7$, $L_{\text{obs}} = 8$.

Suppose $k = 2048$ and $n = 131072$ (128k context).

Therefore, $\frac{M_{\text{ASL}}}{M_{\text{full}}} \approx 0.016$ for both LLMs.

FastKV's memory usage equals to $M_{\text{ASL-KV}}$, thereby yielding approximately the same prediction as above.

Table~\ref{tab:memory} shows that the empirical ratio is $0.018$ for Llama-3.1-8B-UL and $0.027$ for Qwen2.5-7B. 
The discrepancy can be attributed to additional memory consumption such as workspace for FlashAttention, PyTorch memory management, and activations for FFN.

\section{Experimental Setup Details}
\label{sec:exp-setup-detail}

\subsection{Datasets}

\myparagraph{InfiniteBench}~\citep{zhang2024bench}, is a benchmark testing LLMs in various aspects of long-context processing. 
We use the version provided in the MInference GitHub repository~\citep{minference}. 
There are a a total of 3,992 examples, with an average context length of 214k, evaluating the following 10 tasks:
\begin{inparaenum} [(1)]
    \item summarization of a fake book created with core entity substitution (En.Sum a.k.a. longbook\_sum\_eng), 
    \item free-form question answering based on the fake book (En.QA a.k.a. longbook\_qa\_eng), 
    \item multiple choice questions derived from the fake book (En.MC a.k.a. longbook\_choice\_eng), 
    \item identification of talkers in partially anonymized scripts (En.Dia a.k.a. longdialogue\_qa\_eng), 
    \item question answering on a set of Chinese books (Zh.QA a.k.a. longbook\_qa\_chn), 
    \item finding which function in a code repo contains an crashing error (Code.Debug a.k.a. code\_debug), 
    \item finding special integers in a lengthy list (Math.Find a.k.a. math\_find), 
    \item retrieving hidden keys in a noisy long context (Retr.PassKey a.k.a. passkey), 
    \item locating repeated hidden numbers in a noisy long context (Retr.Num a.k.a. number\_string), and
    \item finding the corresponding value from a dictionary and a key (Retr.KV a.k.a. kv\_retrieval), 
\end{inparaenum}

\myparagraph{RULER}~\citep{hsieh2024ruler} is a benchmark of synthetic examples for evaluating long-context LLMs with configurable sequence length and task complexity. 
It consists of 13 tasks, including
\begin{inparaenum} [(1)]
    \item identifying common words from a mixture of common and uncommon words (cwe), 
    \item identifying most frequent words from a Zeta distribution (fwe), 
    \item single-key NIAH, where a single key-value pair is inserted into noisy text, with varying difficulties (niah\_single\_1, niah\_single\_2, and niah\_single\_3 a.k.a. S-NIAH1, S-NIAH2, and S-NIAH3, respectively), 
    \item multi-key NIAH, where multiple keys are inserted and a specific value among hard distractors needs to be retrieved, with varying difficulties (niah\_multikey\_1, niah\_multikey\_2, and niah\_multikey\_3 a.k.a. MK-NIAH1, MK-NIAH2, and MK-NIAH3, respectively), 
    \item retrieving values for multiple keys (niah\_multiquery a.k.a. MQ-NIAH), 
    \item retrieving all values associated with a single key (niah\_multivalue a.k.a. MV-NIAH), 
    \item question answering with distracting paragraphs inserted, with varying difficulties (qa\_1 and qa\_2 a.k.a. QA1 and QA2, respectively), 
    \item tracing all variable names pointing to the same value through variable bindings (vt). 
\end{inparaenum}
We test models on 4k, 8k, 16k, 32k, 64k, and 128k context lengths, including 2,600 examples per length. 

\myparagraph{NIAH}~\citep{niah} is simple needle-in-a-haystack analysis to test in-context retrieval ability of long-context LLMs. The model needs to identify a random fact or statement (the ``needle'') from a long-context window (the ``haystack''). The evaluation iterates over document depths (where the needle is placed) and context lengths. We scale the test from 1k to 256k. 

\subsection{Methods}
We compare our method, ASL, with three layer-wise token pruning methods: 
\begin{itemize} 
  \item GemFilter~\citep{shi2024discovering}, a one-shot method using a two-pass strategy for token selection. In the first pass, it calculates full attention for the first $L$ layers, finding the \topk tokens at the $L$-th layer; in the second pass, it calculates attention for the selected token set for all layers.
  \item FastKV~\citep{jo2025fastkv}, a one-shot method similar to GemFilter but having only one pass. Before reaching the selection layer, full attention is calculated for prefilling but only \topk tokens are retained in the KV cache for decoding. From the selection layer, only \topk selected tokens are carried forward to deeper layers for both prefilling and decoding. 
  \item PyramidInfer~\citep{yang2024pyramidinfer}, a progressive method that performs \topp selection and adopts a decay ratio to determine the value of $p$ at each layer. 
\end{itemize}

The selection layer for FastKV is 15 for Llama-3.1-8B-UL and 14 for Qwen2.5-7B. 
The selection layer for GemFilter is 13 for Llama-3.1-8B-UL and 14 for Qwen2.5-7B. 
For 1D average pooling in SnapKV and ASL, window size = 32 and kernel size = 7. 

LazyLLM~\citep{fu2024lazyllm} and SlimInfer~\citep{long2025sliminfer}, another two layer-wise token pruning that optimize prefilling, are excluded due to unavailable source codes. MInference~\citep{jiang2024minference}, despite optimizing TTFT, is also excluded because it is orthogonal to our method and its block-sparse kernel does not reduce KV cache. 

\subsection{Environments}
The experiments are conducted on an Nvidia H100 GPU with 80GB memory. 
We run the LLMs with the Hugging Face Transformers library~\citep{wolf-etal-2020-transformers}. 
All the methods use FlashAttention-2~\citep{dao2023flashattention}, except PyramidInfer, which uses eager attention~\citep{hfattention}.

\section{Additional Experiments}
\label{sec:exp-additional}

\begin{table*}[t]
    \small
    \centering
    \setlength{\tabcolsep}{1ex}
    \caption{TTFT ($\downarrow$) comparison on InfiniteBench. Ratio to Full KV ($= 1$) is reported.}
    \resizebox{\textwidth}{!}{
    \begin{tabular}{l|cccccccccc|c}
    \toprule
    \textbf{Methods}       & \textbf{En.Sum} & \textbf{En.QA} & \textbf{En.MC} & \textbf{En.Dia} & \textbf{Zh.QA} & \textbf{Code.Debug} & \textbf{Math.Find} & \textbf{Retr.PassKey} & \textbf{Retr.Num} & \textbf{Retr.KV} & \textbf{Avg.} \\
    \midrule
    \midrule
    \multicolumn{12}{c}{\textbf{Llama-3.1-8B-UL, KV Budget = 2048.}} \\
    \midrule
    SnapKV        & 1.00 & 1.01 & 0.99 & 0.99 & 0.99 & 1.01 & 1.01 & 1.02 & 1.01 & 1.01 & 1.00 \\
    FastKV        & 0.50 & 0.50 & 0.51 & 0.50 & 0.51 & 0.50 & 0.50 & 0.52 & 0.50 & 0.50 & 0.50 \\
    \rowcolor[HTML]{DAE8FC}
    ASL           & 0.58 & 0.73 & 0.76 & 0.71 & 7.70 & 0.67 & 0.67 & 0.47 & 0.60 & 0.68 & 0.68 \\
    GemFilter     & 0.44 & 0.44 & 0.44 & 0.43 & 0.44 & 0.45 & 0.44 & 0.45 & 0.44 & 0.44 & \textbf{0.44} \\
    \rowcolor[HTML]{DAE8FC}
    \aslgf        & 0.58 & 0.74 & 0.77 & 0.71 & 0.71 & 0.69 & 0.67 & 0.48 & 0.62 & 0.68 & 0.69 \\
    \midrule
    \midrule
    \multicolumn{12}{c}{\textbf{Qwen2.5-7B, KV Budget = 2048.}} \\
    \midrule
    SnapKV        & 1.00 & 1.00 & 1.00 & 1.00 & 1.00 & 1.00 & 1.00 & 1.00 & 1.00 & 1.00 & 1.00 \\
    FastKV        & 0.54 & 0.53 & 0.54 & 0.53 & 0.54 & 0.54 & 0.54 & 0.54 & 0.53 & 0.53 & \textbf{0.53} \\
    \rowcolor[HTML]{DAE8FC}
    ASL           & 0.80 & 0.82 & 0.90 & 0.85 & 0.85 & 0.82 & 0.73 & 0.77 & 0.70 & 0.73 & 0.82 \\
    GemFilter     & 0.54 & 0.53 & 0.53 & 0.53 & 0.53 & 0.53 & 0.53 & 0.54 & 0.52 & 0.53 & \textbf{0.53} \\
    \rowcolor[HTML]{DAE8FC}
    \aslgf        & 0.80 & 0.83 & 0.90 & 0.85 & 0.84 & 0.83 & 0.72 & 0.76 & 0.70 & 0.73 & 0.82 \\
    \bottomrule
    \end{tabular}
    }
    \label{tab:ttft-infinitebench}
\end{table*}
\begin{table*}[t]
    \small
    \centering
    \setlength{\tabcolsep}{1ex}
    \caption{TPOT ($\downarrow$) comparison on InfiniteBench. Ratio to Full KV ($= 1$) is reported.}
    \resizebox{\textwidth}{!}{
    \begin{tabular}{l|cccccccccc|c}
    \toprule
    \textbf{Methods}       & \textbf{En.Sum} & \textbf{En.QA} & \textbf{En.MC} & \textbf{En.Dia} & \textbf{Zh.QA} & \textbf{Code.Debug} & \textbf{Math.Find} & \textbf{Retr.PassKey} & \textbf{Retr.Num} & \textbf{Retr.KV} & \textbf{Avg.} \\
    \midrule
    \midrule
    \multicolumn{12}{c}{\textbf{Llama-3.1-8B-UL, KV Budget = 2048.}} \\
    \midrule
    SnapKV        & 0.31 & 0.35 & 0.26 & 0.20 & 0.26 & 0.29 & 0.35 & 0.34 & 0.30 & 0.29 & 0.29 \\
    FastKV        & 0.27 & 0.31 & 0.23 & 0.17 & 0.23 & 0.25 & 0.32 & 0.29 & 0.27 & 0.25 & 0.25 \\
    \rowcolor[HTML]{DAE8FC}
    ASL           & 0.29 & 0.29 & 0.24 & 0.18 & 0.23 & 0.25 & 0.32 & 0.29 & 0.26 & 0.25 & 0.25 \\
    GemFilter     & 0.32 & 0.30 & 0.21 & 0.17 & 0.23 & 0.24 & 0.25 & 0.22 & 0.25 & 0.21 & \textbf{0.23} \\
    \rowcolor[HTML]{DAE8FC}
    \aslgf        & 0.32 & 0.29 & 0.20 & 0.17 & 0.23 & 0.24 & 0.24 & 0.23 & 0.24 & 0.21 & \textbf{0.23} \\
    \midrule
    \midrule
    \multicolumn{12}{c}{\textbf{Qwen2.5-7B, KV Budget = 2048.}} \\
    \midrule
    SnapKV        & 0.16 & 0.12 & 0.12 & 0.09 & 0.12 & 0.13 & 0.18 & 0.15 & 0.15 & 0.15 & 0.13 \\
    FastKV        & 0.16 & 0.12 & 0.12 & 0.09 & 0.12 & 0.13 & 0.18 & 0.15 & 0.15 & 0.15 & 0.13 \\
    \rowcolor[HTML]{DAE8FC}
    ASL           & 0.16 & 0.12 & 0.12 & 0.09 & 0.12 & 0.12 & 0.18 & 0.15 & 0.15 & 0.15 & 0.13 \\
    GemFilter     & 0.15 & 0.09 & 0.09 & 0.07 & 0.09 & 0.09 & 0.13 & 0.07 & 0.11 & 0.11 & \textbf{0.10} \\
    \rowcolor[HTML]{DAE8FC}
    \aslgf        & 0.14 & 0.09 & 0.09 & 0.07 & 0.09 & 0.09 & 0.14 & 0.06 & 0.11 & 0.12 & \textbf{0.10} \\
    \bottomrule
    \end{tabular}
    }
    \label{tab:tpot-infinitebench}
\end{table*}
\begin{figure*}[t]
    \small
    \centering
    \begin{subfigure}{0.33\textwidth}
        \centering
        \includegraphics[width=\linewidth]{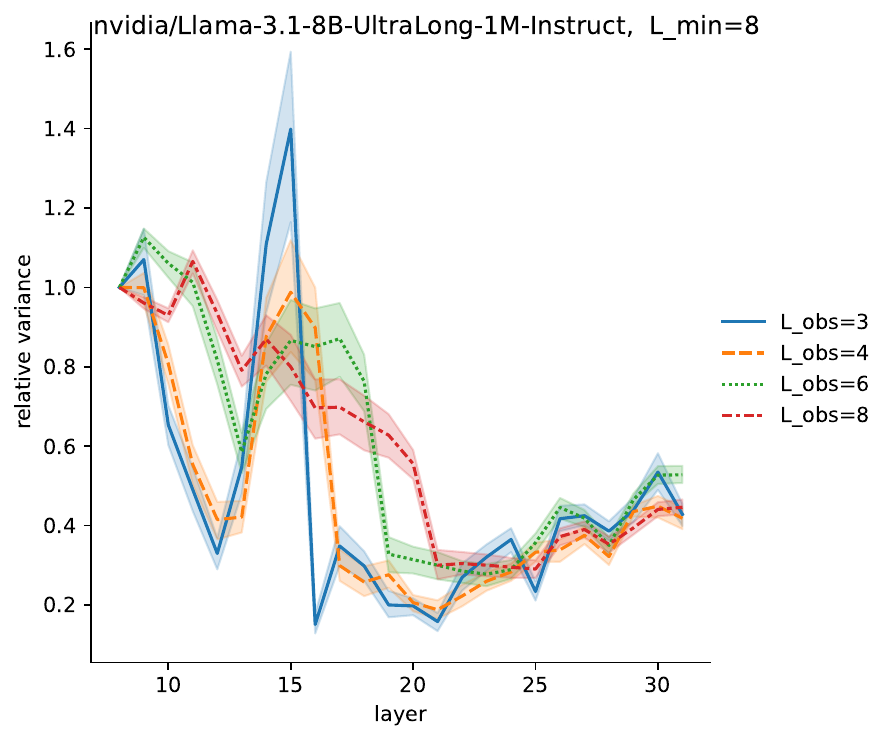}
        \caption{Llama-3.1-8B-UL, $L_{\min} = 8$.}
        \label{fig:ablation-obs-llama318b-lmin8}
    \end{subfigure}
    \begin{subfigure}{0.33\textwidth}
        \centering
        \includegraphics[width=\linewidth]{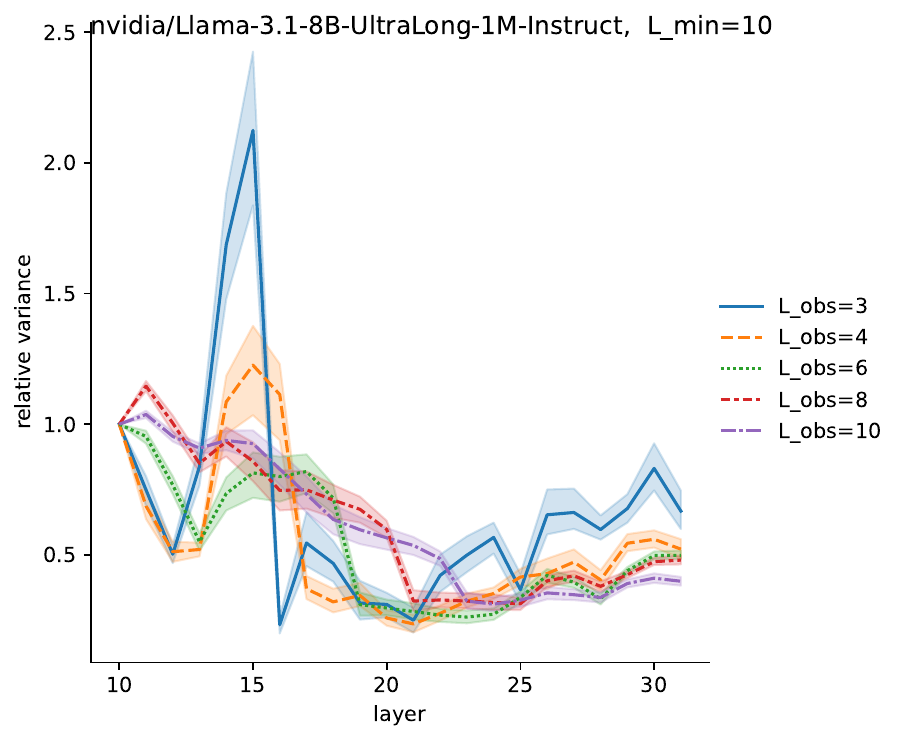}
        \caption{Llama-3.1-8B-UL, $L_{\min} = 10$.}
        \label{fig:ablation-obs-llama318b-lmin10}
    \end{subfigure}
    \begin{subfigure}{0.33\textwidth}
        \centering
        \includegraphics[width=\linewidth]{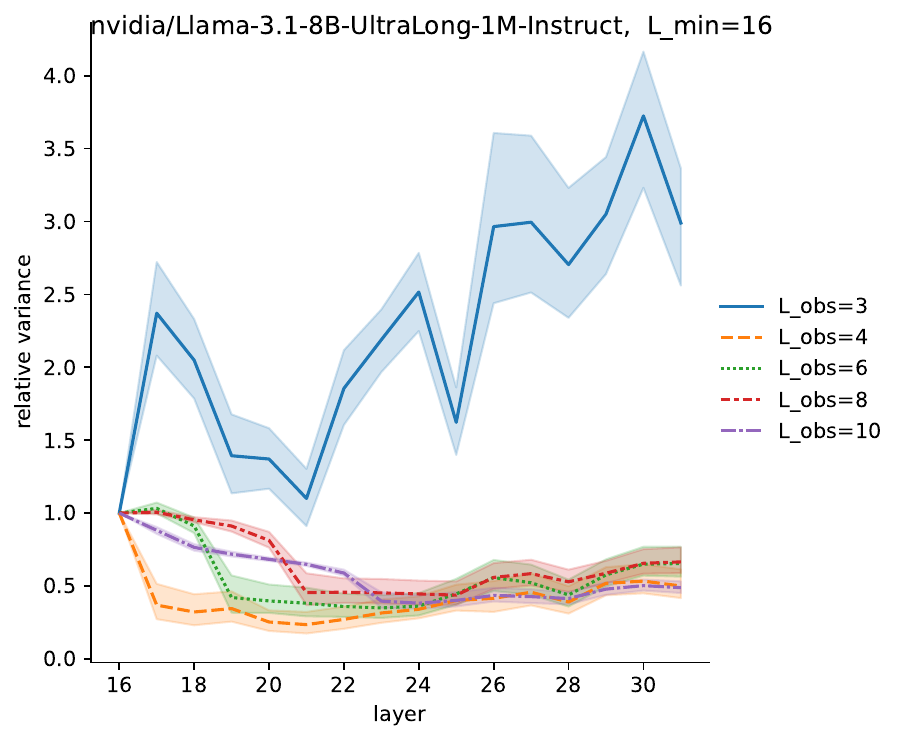}
        \caption{Llama-3.1-8B-UL, $L_{\min} = 16$.}
        \label{fig:ablation-obs-llama318b-lmin16}
    \end{subfigure}
    \begin{subfigure}{0.33\textwidth}
        \centering
        \includegraphics[width=\linewidth]{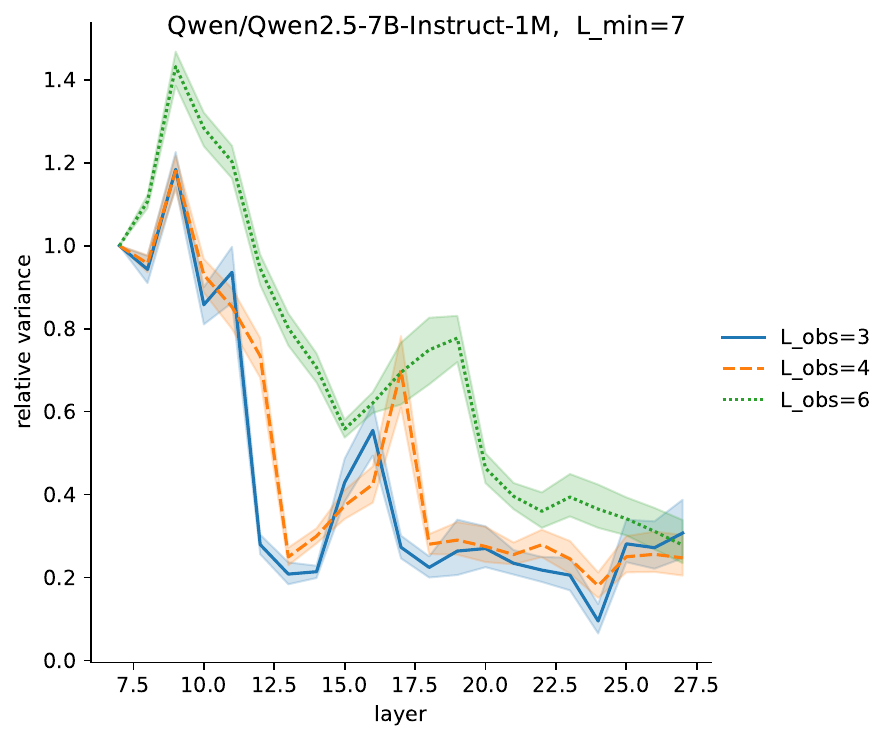}
        \caption{Qwen2.5-7B, $L_{\min} = 7$.}
        \label{fig:ablation-obs-qwen257b-lmin7}
    \end{subfigure}
    \begin{subfigure}{0.33\textwidth}
        \centering
        \includegraphics[width=\linewidth]{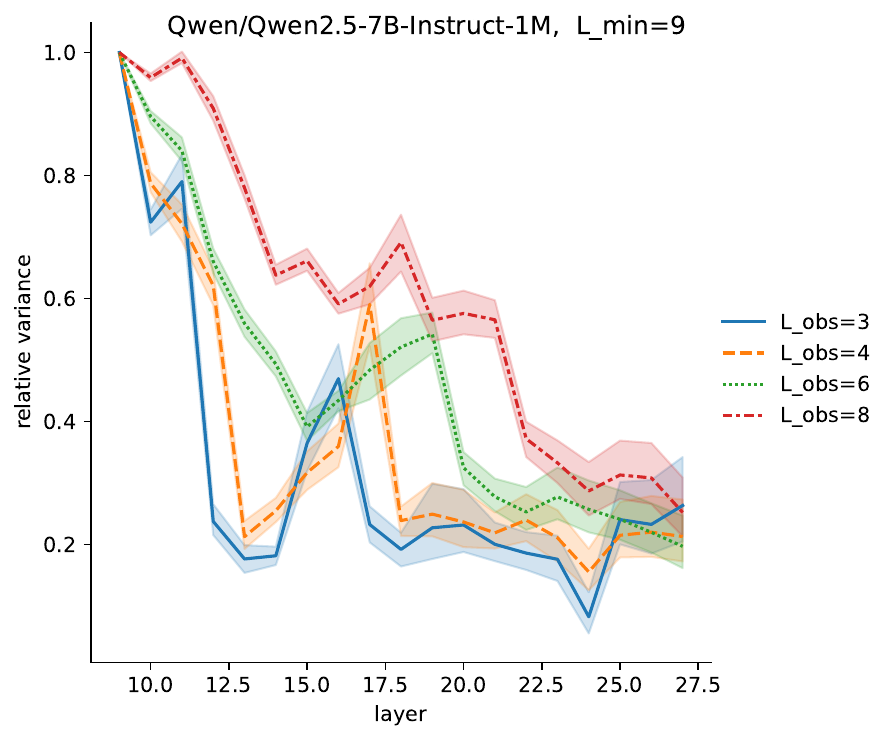}
        \caption{Qwen2.5-7B, $L_{\min} = 9$.}
        \label{fig:ablation-obs-qwen257b-lmin9}
    \end{subfigure}
    \begin{subfigure}{0.33\textwidth}
        \centering
        \includegraphics[width=\linewidth]{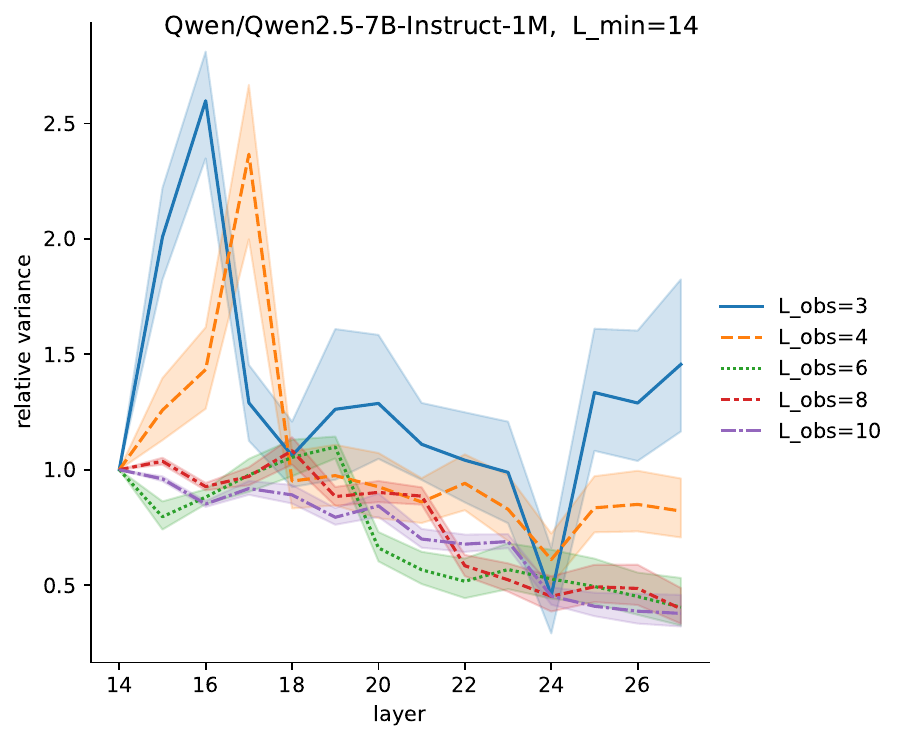}
        \caption{Qwen2.5-7B, $L_{\min} = 14$.}
        \label{fig:ablation-obs-qwen257b-lmin14}
    \end{subfigure}
    \caption{Effect of $L_{\min}$ and $L_{\obs}$ on relative variance, KV retrieval task.}
    \label{fig:ablation-obs}
\end{figure*}

\subsection{TTFT and TPOT on InfiniteBench}
Tables~\ref{tab:ttft-infinitebench} and~\ref{tab:tpot-infinitebench} report the TTFT and TPOT on the 10 tasks of InfiniteBench. 
Compared to the results on RULER (Tables~\ref{tab:ttft-ruler} and~\ref{tab:tpot-ruler}), the observations are similar. 
ASL and \aslgf trade prefilling time for higher accuracy, while the decoding speed is approximately the same as other layer-wise methods.

\subsection{Effect of $L_{\min}$ and $L_{\obs}$}
To study the effect of observation start layer number $L_{\min}$ and lookback layer number $L_{\obs}$, we plot the relative variance under different settings of $L_{\min}$ and $L_{\obs}$ in Figure~\ref{fig:ablation-obs}. 

\myparagraph{$L_{\min}$}
From Figure~\ref{fig:ablation-obs}, we observe that the relative variance generally exhibits a decreasing trend under smaller values of $L_{\min}$. 
In contrast, the relative variance fluctuates more violently and even rebounds under larger $L_{\min}$ values.
In order to find a threshold to determine the selection layer, we choose to use $L_{\min} = \floor{L_{\text{model}}/3}$ in ASL, where $L_{\text{model}}$ is the number of the layers in the LLM, e.g., $\floor{32/3} = 10$ for Llama-3.1-8B-UL and $\floor{28/3} = 9$ for Qwen2.5-7B. 
This is to strike a balance between the identification of the steepest decline in variance and the additional overhead (because we only start the operation of ASL when reaching layer $L_{\min}$). 

\myparagraph{$L_{\obs}$}
As shown in Figure~\ref{fig:ablation-obs}, $L_{\obs}$ represents the sensitivity of the relative variance to attention scores, which change across layers. A smaller $L_{\obs}$ suggests that the variance is more responsive to the change of layers. 
Seeing this, we choose $L_{\obs}$ to be 8 control the sensitivity and effectiveness (i.e., the decreasing trend can be identified, so we can use a threshold to determine the selection layer). 

\subsection{Effect of Relative Variance Threshold}
\label{sec:exp-effect-tau-tasks}


We report the effect of $\tau$ across the 13 tasks of RULER in Table~\ref{tab:various-tau-ruler}, where accuracy is measured at a context length of 128k. 
When the KV Budget is set to 2048, the general trend is that accuracy varies only slightly across different $\tau$ settings for all 13 tasks, except for outliers observed on niah\_single\_3 with Qwen2.5-7B at the smallest ($\tau=0.2$) and largest ($\tau=0.6$) values.
In contrast, when KV compression methods are not applied (i.e., when the KV Budget is full before the selection layer), the accuracy on tasks such as niah\_single\_3 and niah\_multikeys\_3 are highly sensitive to $\tau$.
For example, with Qwen2.5-7B on niah\_multikeys, the accuracy exceeds 70 at $\tau = 0.3$ but drops sharply to only 3.0 at $\tau = 0.6$. 

\begin{table*}[t]
    \small
    \centering
    \setlength{\tabcolsep}{1ex}
    \caption{Effect of relative variance threshold $\tau$ on accuracy ($\uparrow$), RULER, 128k.}
    \resizebox{\textwidth}{!}{
    \begin{tabular}{l|ccccccccccccc|c}
    \toprule
    \textbf{$\tau$}  & \textbf{S-NIAH1} & \textbf{S-NIAH2} & \textbf{S-NIAH3} & \textbf{MK-NIAH1} & \textbf{MK-NIAH2} & \textbf{MK-NIAH3} & \textbf{MV-NIAH} & \textbf{MQ-NIAH}& \textbf{VT}& \textbf{CWE}& \textbf{FWE}&\textbf{QA1} &\textbf{QA2}&\textbf{Avg.}\\
    \midrule
    \midrule
    \multicolumn{15}{c}{\textbf{Llama-3.1-8B-UL, KV Budget = 2048.}} \\
    \midrule
    0.2        & 100.0 & 98.0 & 63.0 & 89.5 & 15.0 & 0.5 & 46.6 & 74.4 & 75.2 & 0.15 & 67.2 & 58.0 & 34.5 & 55.5 \\
    0.3        & 100.0 & 98.0 & 44.0 & 88.5 & 15.0 & 0.5 & 72.6 & 73.9 & 75.3 & 0.05 & 67.0 & 60.5 & 34.5 & 56.1 \\
    0.4        & 100.0 & 95.0 & 45.0 & 87.5 & 7.5 & 0.5 & 74.9 & 71.9 & 75.6 & 0.05 & 65.7 & 58.0 & 34.0 & 55.0 \\
    0.5        & 100.0 & 96.0 & 64.5 & 90.0 & 10.0 & 0.5 & 76.9 & 70.8 & 75.4 & 0.05 & 65.2 & 58.5 & 35.0 & 57.1 \\
    0.6        & 100.0 & 96.0 & 71.0 & 89.0 & 10.5 & 0.5 & 76.6 & 75.0 & 76.1 & 0.05 & 65.2 & 58.5 & 34.0 & 57.9 \\
    \midrule
    \midrule
    \multicolumn{15}{c}{\textbf{Llama-3.1-8B-UL, KV Budget = Full (before selection) \& 2048 (after selection).}} \\
    \midrule
    0.2        & 100.0 & 98.5 & 99.0 & 90.0 & 68.5 & 15.0 & 67.6 & 87.6 & 83.4 & 0.05 & 90.5 & 56.0 & 35.0 & 68.6 \\
    0.3        & 100.0 & 98.5 & 90.0 & 89.5 & 68.0 & 14.5 & 87.5 & 87.4 & 83.1 & 0.05 & 87.3 & 60.0 & 33.5 & 69.2 \\
    0.4        & 100.0 & 95.5 & 63.0 & 88.5 & 31.5 & 11.5 & 91.5 & 83.6 & 83.0 & 0.05 & 82.0 & 58.0 & 33.0 & 63.2 \\
    0.5        & 100.0 & 97.0 & 72.5 & 90.0 & 26.5 & 10.5 & 90.9 & 83.0 & 83.1 & 0.05 & 81.0 & 58.0 & 34.0 & 63.6 \\
    0.6        & 100.0 & 96.5 & 76.0 & 89.0 & 29.5 & 8.0 & 91.3 & 85.4 & 82.7 & 0.05 & 79.3 & 58.5 & 33.5 & 63.8 \\
    \midrule
    \midrule
    \multicolumn{15}{c}{\textbf{Qwen2.5-7B, KV Budget = 2048.}} \\
    \midrule
    0.2        & 100.0 & 99.5 & 16.5 & 99.5 & 85.0 & 0.0 & 82.9 & 98.9 & 89.1 & 10.1 & 56.5 & 61.0 & 44.7 & 64.9 \\
    0.3        & 100.0 & 99.5 & 33.5 & 99.5 & 85.0 & 1.5 & 83.4 & 98.9 & 88.1 & 11.1 & 60.7 & 58.0 & 44.7 & 66.4 \\
    0.4        & 100.0 & 99.5 & 36.0 & 100.0 & 85.0 & 1.0 & 83.0 & 99.0 & 88.0 & 11.4 & 61.2 & 61.0 & 44.7 & 66.9 \\
    0.5        & 100.0 & 97.5 & 23.5 & 99.0 & 85.5 & 1.0 & 79.9 & 98.8 & 88.2 & 11.4 & 62.5 & 62.0 & 45.5 & 65.7 \\
    0.6        & 100.0 & 88.5 & 7.5 & 92.5 & 86.0 & 1.0 & 70.0 & 97.5 & 88.5 & 11.4 & 62.2 & 59.0 & 43.2 & 62.1 \\

    \midrule
    \midrule
    \multicolumn{15}{c}{\textbf{Qwen2.5-7B, KV Budget = Full (before selection) \& 2048 (after selection).}} \\
    \midrule
    0.2        & 100.0 & 99.5 & 84.5 & 99.5 & 96.5 & 87.0 & 85.5 & 99.6 & 83.1 & 9.3 & 62.3 & 61.5 & 47.7 & 78.2 \\
    0.3        & 100.0 & 99.5 & 48.0 & 99.5 & 96.5 & 72.5 & 85.5 & 99.6 & 81.9 & 11.5 & 62.3 & 59.3 & 48.2 & 74.2 \\
    0.4        & 100.0 & 99.5 & 43.0 & 100.0 & 96.5 & 26.5 & 86.0 & 99.6 & 81.6 & 12.1 & 62.7 & 62.3 & 46.2 & 70.5 \\
    0.5        & 100.0 & 97.0 & 27.5 & 99.5 & 96.5 & 9.0 & 83.0 & 99.6 & 82.7 & 12.1 & 62.5 & 62.0 & 45.5 & 67.5 \\
    0.6        & 100.0 & 88.5 & 8.5 & 93.5 & 97.0 & 3.0 & 74.3 & 98.3 & 84.2 & 12.0 & 61.3 & 59.3 & 43.2 & 63.3 \\
    
    \bottomrule
    \end{tabular}
    }
    \label{tab:various-tau-ruler}
\end{table*}


\subsection{Distributions of Selection Layer}
\label{sec:exp-selection-layer-dist}

\begin{figure*}[t]
    \centering
    \includegraphics[width=\linewidth]{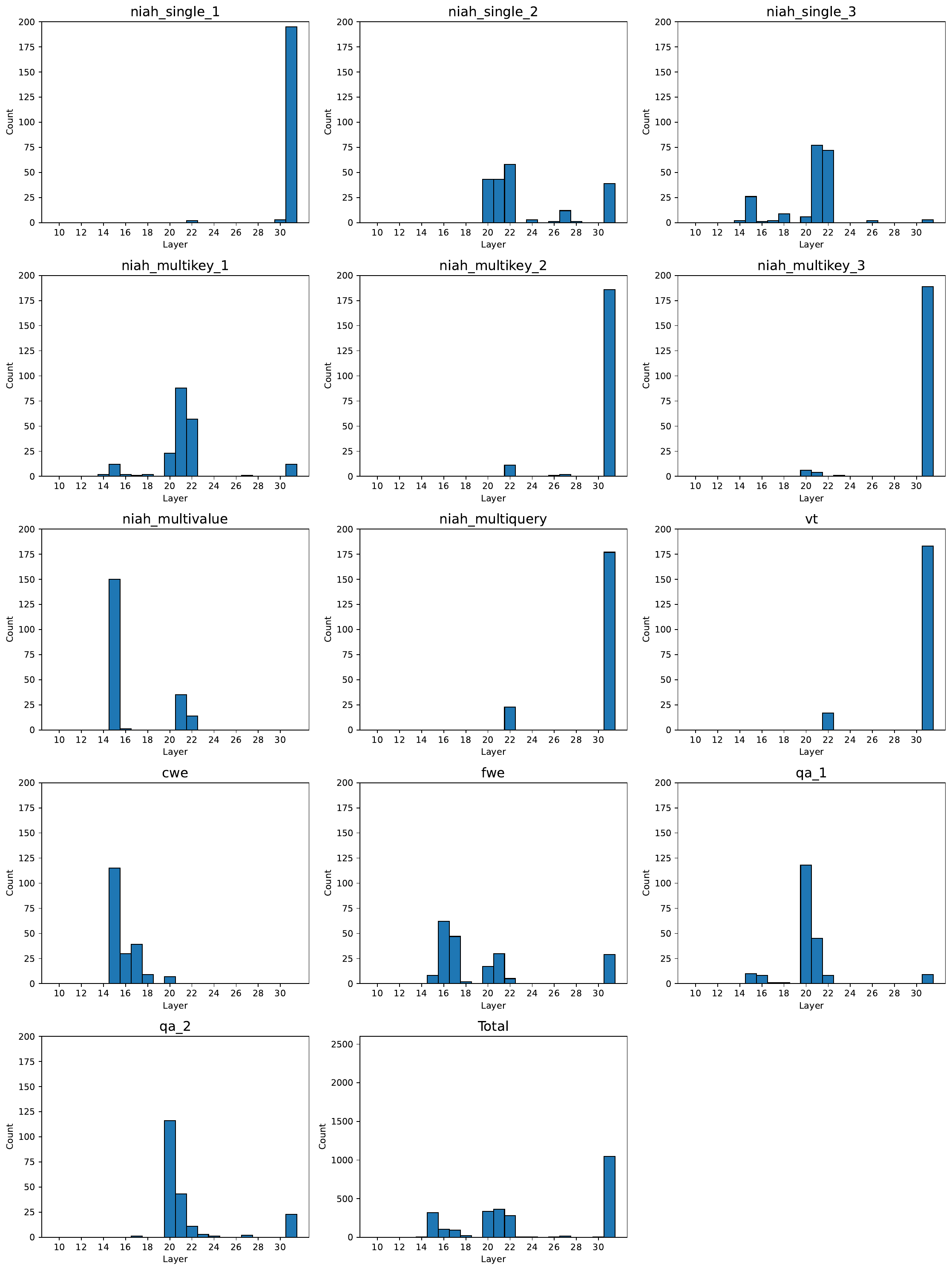}
    \caption{Distributions of selection layer across tasks, Llama-3.1-8B-UL, RULER, 128k, $\tau = 0.3$.}
    \label{fig:Distribution_of_select_layer_Llama-3.1-8B-UltraLong-1M-Instruct}
\end{figure*}

\begin{figure*}[t]
    \centering
    \includegraphics[width=\linewidth]{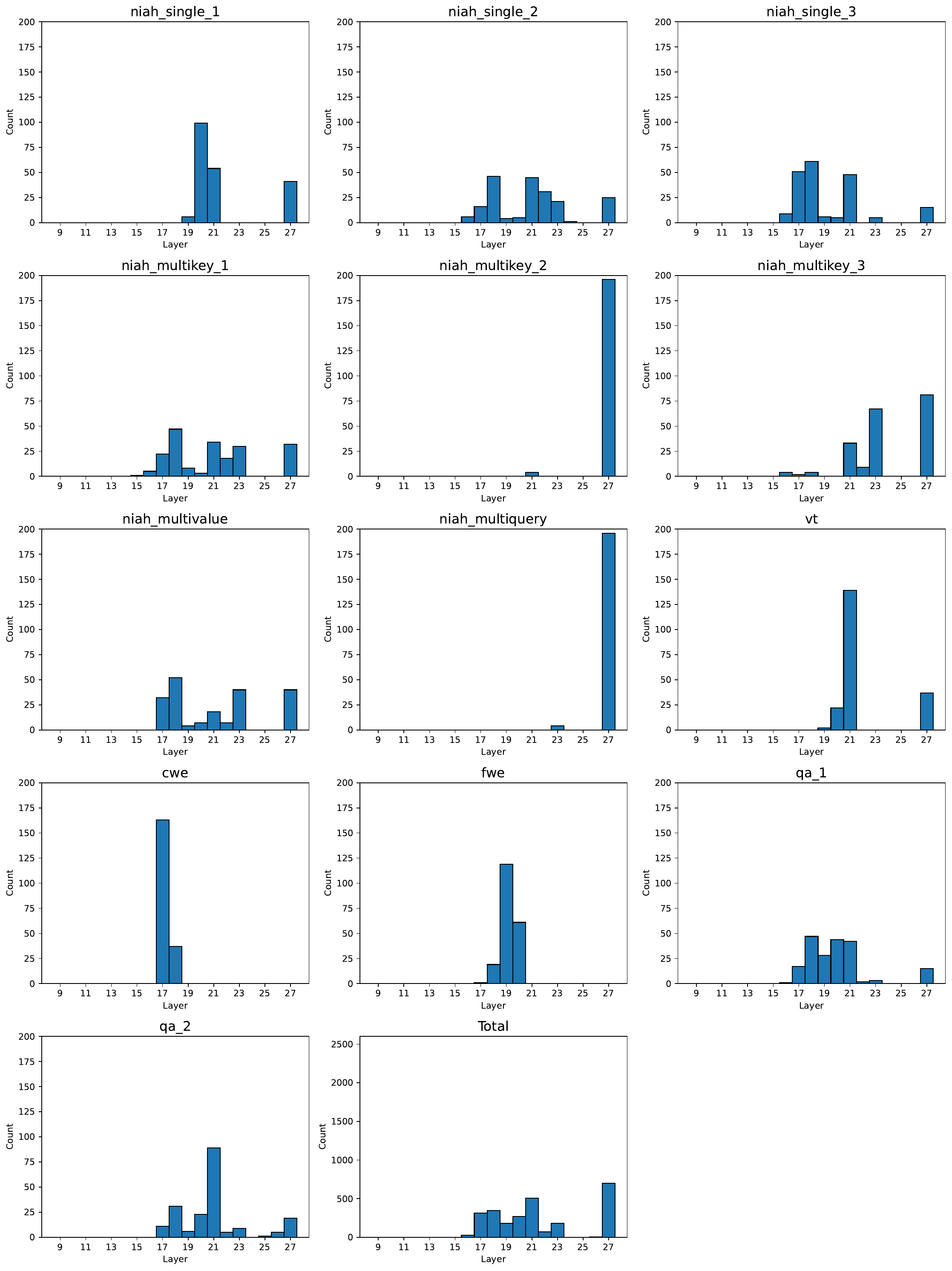}
    \caption{Distributions of selection layer across tasks, Qwen2.5-7B, RULER, 128k, $\tau = 0.3$.}
    \label{fig:Distribution_of_select_layer_Qwen2.5-7B-Instruct-1M}
\end{figure*}

Figures~\ref{fig:Distribution_of_select_layer_Llama-3.1-8B-UltraLong-1M-Instruct} and~\ref{fig:Distribution_of_select_layer_Qwen2.5-7B-Instruct-1M} show the frequency of a layer determined as the selection layer RULER, 128k context, and a KV budget of 2048. Each task in RULER consists of 200 questions. Compared to FastKV, which uses fixed layer selection, we observe that in tasks where FastKV reports lower scores, ASL selects a later layer than FastKV, while in tasks where FastKV reports higher scores, ASL picks a selection layer close to FastKV's.

\newpage

\end{document}